\documentclass[lettersize,journal]{IEEEtran}
\usepackage{amsmath,amsfonts}
\usepackage{algorithmic} 
\usepackage{algorithm}
\usepackage{array}
\usepackage{amsmath}
\usepackage[caption=false,font=footnotesize,labelfont=sf,textfont=sf]{subfig}
\usepackage{textcomp}
\usepackage{stfloats}
\usepackage{url}
\usepackage{verbatim}
\usepackage{graphicx}
\usepackage{cite}
\usepackage{adjustbox} 
\usepackage{booktabs}   
\usepackage{multirow}
\usepackage{makecell}
\usepackage{xcolor}
\usepackage[normalem]{ulem}
\usepackage{comment}

\begin{document}

\title{LiftProj: Space Lifting and Projection-Based Panorama Stitching}

\author{Yuan Jia, Ruimin Wu, Rui Song, \IEEEmembership{Member,~IEEE}, Jiaojiao Li, \IEEEmembership{Senior Member,~IEEE}, and Bin Song, \IEEEmembership{Senior Member,~IEEE}
\thanks{This work was supported in part by the National Nature Science Foundation of China under Grant xxxxxx}
\thanks{Manuscript received December 29, 2025; revised January 16, 2026.}}

\markboth{Journal of \LaTeX\ Class Files,~Vol.~14, No.~8, August~2021}%
{Shell \MakeLowercase{\textit{et al.}}: A Sample Article Using IEEEtran.cls for IEEE Journals}

\IEEEpubid{0000--0000/00\$00.00~\copyright~2021 IEEE}

\maketitle

\begin{abstract}
Traditional image stitching techniques have predominantly utilized two-dimensional homography transformations and mesh warping to achieve alignment on a planar surface. While effective for scenes that are approximately coplanar or exhibit minimal parallax, these approaches often result in ghosting, structural bending, and stretching distortions in non-overlapping regions when applied to real three-dimensional scenes characterized by multiple depth layers and occlusions. Such challenges are exacerbated in multi-view accumulations and 360° closed-loop stitching scenarios. In response, this study introduces a spatially lifted panoramic stitching framework that initially elevates each input image into a dense three-dimensional point representation within a unified coordinate system, facilitating global cross-view fusion augmented by confidence metrics. Subsequently, a unified projection center is established in three-dimensional space, and an equidistant cylindrical projection is employed to map the fused data onto a single panoramic manifold, thereby producing a geometrically consistent 360° panoramic layout. Finally, hole filling is conducted within the canvas domain to address unknown regions revealed by viewpoint transitions, restoring continuous texture and semantic coherence. This framework reconceptualizes stitching from a two-dimensional warping paradigm to a three-dimensional consistency paradigm and is designed to flexibly incorporate various three-dimensional lifting and completion modules. Experimental evaluations demonstrate that the proposed method substantially mitigates geometric distortions and ghosting artifacts in scenarios involving significant parallax and complex occlusions, yielding panoramic results that are more natural and consistent.
\end{abstract}

\begin{IEEEkeywords}
Article submission, IEEE, IEEEtran, journal, \LaTeX, paper, template, typesetting.
\end{IEEEkeywords}

\section{Introduction}
\IEEEPARstart{I}{mage}  stitching constitutes a fundamental task within the domain of computer vision, with extensive applications in areas such as security surveillance, intelligent driving, autonomous robotic navigation, and immersive content creation. Notably, panoramic stitching for 360° surround imaging enables the generation of a continuous field-of-view representation from a single viewpoint, which is critically important for applications including spatial perception and scene reconstruction \cite{Kang_1998}. Nevertheless, real-world environments frequently present significant three-dimensional structural variations and occlusion relationships. Challenges such as pronounced parallax, locally sparse texture, and extensive non-overlapping regions between input viewpoints complicate the production of high-quality, distortion-free panoramas.

Conventional mainstream approaches predominantly perform image registration and warping on a two-dimensional plane using homography matrices or their extensions. To enhance alignment accuracy within overlapping regions, prior research has introduced local models based on multiple homographies or mesh warping techniques, alongside constraints aimed at preserving lines and structural integrity to mitigate distortion \cite{2014Sphp,2015Aanap,2013Apap,2016gsp,2022ges,2021lpc}. While these methods can improve visual quality in certain contexts, they fundamentally approximate three-dimensional parallax within a two-dimensional framework. Consequently, pixels corresponding to different depth layers are compelled to undergo identical types of two-dimensional transformations, resulting in inevitable ghosting artifacts and structural deformation. Furthermore, distortions tend to propagate into non-overlapping regions during multi-view accumulation. Recent deep learning-based stitching methods have enhanced matching robustness through end-to-end learning paradigms \cite{2018deephomo,2023udis++,2020Content,2023pixel,2025PixelStitch}, yet they remain reliant on the two-dimensional warping paradigm. As a result, these approaches face difficulties in explicitly modeling three-dimensional correspondences and occlusion relationships across views at the geometric level, leading to frequent global structural inconsistencies, particularly under conditions of strong parallax and in 360° closed-loop scenarios.

A critical insight is that these challenges do not primarily arise from insufficient feature representation or optimization strategies, but rather from the intrinsic limitations of the two-dimensional stitching paradigm, which cannot simultaneously ensure multi-depth geometric consistency and global projection coherence. Recent advances in dense three-dimensional reconstruction techniques that obviate the need for camera calibration—such as DUSt3R \cite{2024dust3r} and VGGT \cite{wang2025vggt}—have made it feasible to directly recover dense 3D correspondences and relative camera poses from multi-view images. This progress enables the elevation of the stitching process into three-dimensional space, facilitating global alignment within a unified geometric framework.

Motivated by these developments, the present study proposes a spatially elevated panoramic stitching framework. Initially, each input image is transformed into a dense three-dimensional point representation within a unified coordinate system, with cross-view fusion guided by confidence metrics. Subsequently, a unified projection center is established in three-dimensional space, and the fused point cloud is projected onto a single panoramic manifold via equidistant cylindrical projection, thereby producing a geometrically consistent 360° panoramic layout. Finally, hole filling is conducted within the canvas domain to address unknown regions revealed after viewpoint transitions, resulting in continuous and natural panoramic outputs. It is important to note that the 3D lifting and completion modules are presented as interchangeable implementations; the principal contribution of this work lies in the overall framework design that enables 3D-consistent stitching and unified projection.

The principal contributions of this research are as follows: (1) the introduction of a panoramic stitching framework that transitions from the traditional two-dimensional warping paradigm to a three-dimensional consistency paradigm, achieving cross-view geometric alignment and fusion within three-dimensional space; (2) the development of an equidistant cylindrical projection strategy with a unified projection center to realize globally consistent projection for 360° surround views, substantially mitigating multi-view accumulated distortion; and (3) the integration of a hole-filling procedure coupled with three-dimensional projection in the canvas domain to address unobservable regions caused by viewpoint switching. Experimental evaluations demonstrate the effectiveness and robustness of the proposed framework under conditions of strong parallax and complex occlusions.

The remainder of this paper is organized as follows: Section 2 provides a review of the technical evolution of related work; Section 3 elaborates on the spatially elevated stitching framework and implementation details; Section 4 presents the experimental setup alongside quantitative and qualitative comparative results.

\section{Related Work}

From the initial global transformation frameworks relying on handcrafted features to contemporary approaches driven by deep learning, research in image stitching has persistently concentrated on two principal challenges: enhancing feature matching accuracy and mitigating geometric distortion. This section provides a comprehensive review of pertinent studies from both traditional and deep learning perspectives, critically examining the limitations of existing methodologies in addressing issues related to multi-depth object alignment and projection distortion.

\subsection{Traditional Feature-Based Image Stitching Methods}

\textbf{Enhancing Registration Accuracy.} Traditional image registration techniques are founded upon the theoretical principles of scale-invariant feature detection and global geometric transformations. Notably, AutoStitch \cite{2007Automatic} pioneered the robust implementation of global transformations by leveraging the scale-invariant feature transform (SIFT) \cite{2004Distinctive} alongside homography matrices. Subsequent research efforts have predominantly aimed at refining and optimizing this foundational framework. Given the planar constraints inherent to homography matrices, various adaptive deformation models have been proposed to improve registration accuracy by assigning distinct deformation parameters to planes situated at different depths. The DHW model \cite{2011DHW} introduced an early prototype of multi-homography transformation. Building upon this, APAP \cite{2013Apap} achieved local adaptive homography adjustments through mesh partitioning, thereby establishing the foundational "global + local" feature registration paradigm. Further developments, including Ela \cite{2018ela}, TFA \cite{2019tfa}, and WRBIS \cite{2020Wrbis}, have proposed diverse adaptive deformation models within this framework. Additionally, several studies \cite{2021lpc,2015dfw,2019spw,2017Image}, have integrated both point and line features to estimate transformation matrices from multiple dimensions, particularly when point features are sparse, resulting in significant improvements in stitching quality. While these algorithms, by constructing more flexible deformation parameter spaces, have substantially enhanced registration performance, they frequently introduce unnatural distortions.

\textbf{Mitigating Geometric Distortion.} To generate panoramas with more natural appearance, the scholarly community has primarily addressed structural preservation from both single-angle and global optimization perspectives \cite{2025PixelStitch}. Single-angle optimization techniques, such as QH \cite{2018Quasi}, SPW \cite{2019spw}, and LPC \cite{2021lpc}, incorporate additional terms into the energy function to quantify distortion levels. Conversely, global optimization methods, including SPHP \cite{2014Sphp}, AANAP \cite{2015Aanap}, and GSP \cite{2016gsp}, exploit global similarity constraints to prevent distortion in non-overlapping regions and facilitate smooth transitions. Recently, an increasing number of feature consistency approaches have been integrated into stitching frameworks to enhance structural fidelity. These range from initially extracted line segment features via LSD \cite{2010LSD,2015dfw} to large-scale edges \cite{2022ges}, depth maps \cite{2022depmap}, semantic planes \cite{2021semplaner}, and SAM features \cite{2024Objgsp}. However, the complexity involved in designing these traditional handcrafted features limits the generalizability of stitching methods in complex scenes. Moreover, maintaining structural integrity in non-overlapping regions often necessitates compromises in alignment accuracy.

\subsection{Deep Learning-Based Image Stitching Methods}

In contrast to traditional feature-based methods, deep learning approaches autonomously learn semantic features from data and enable end-to-end stitching solutions, exhibiting robust performance even in low-texture environments. Since the advent of HomographyNet, the first deep network designed to estimate homographies, numerous subsequent methods have extended this network architecture. Nie et al. \cite{2020udisd,2021nie} pioneered the construction of synthetic stitching datasets and proposed an end-to-end benchmark model adhering to the traditional stitching pipeline. Reconstruction-based approaches \cite{2024bfr,2021udis} address misalignment by reconstructing stitched images from semantic features, thereby overcoming the challenge of missing data labels in real-world scenarios. Differentiable elastic deformation algorithms \cite{2024dbc,2023udis++,2024RecStitchNet} facilitate flexible mesh-level deformations. Further advancements include pixel-level deformation schemes \cite{2023pixel,2023pww,2025PixelStitch} that achieve higher alignment precision. Presently, deep learning methods predominantly focus on enhancing alignment accuracy within overlapping regions but often overlook the generation of panoramas with natural appearance. Consequently, their outputs frequently exhibit pronounced distortions, which can lead to failures in multi-image stitching tasks.

Traditional methods, constrained by handcrafted features and explicit optimization capabilities, often require a trade-off between alignment accuracy and structural preservation. Although deep learning techniques substantially improve feature robustness and registration stability, most remain confined to two-dimensional deformation and projection frameworks. These methods can only indirectly compensate for three-dimensional parallax and occlusion, rendering it challenging to ensure global consistency in multi-view and loop-closure scenarios from a geometric standpoint. Concurrently, recent progress in uncalibrated dense 3D reconstruction models presents new opportunities to elevate stitching problems into three-dimensional space. In this context, the present work explores a spatial dimension-raising stitching paradigm that systematically suppresses the fundamental artifacts associated with traditional two-dimensional methods through three-dimensional fusion and panoramic mapping employing a unified projection center.

\section{Proposed Algorithm}

\begin{figure*}[htbp]
    \centering
    \includegraphics[width=1\linewidth]{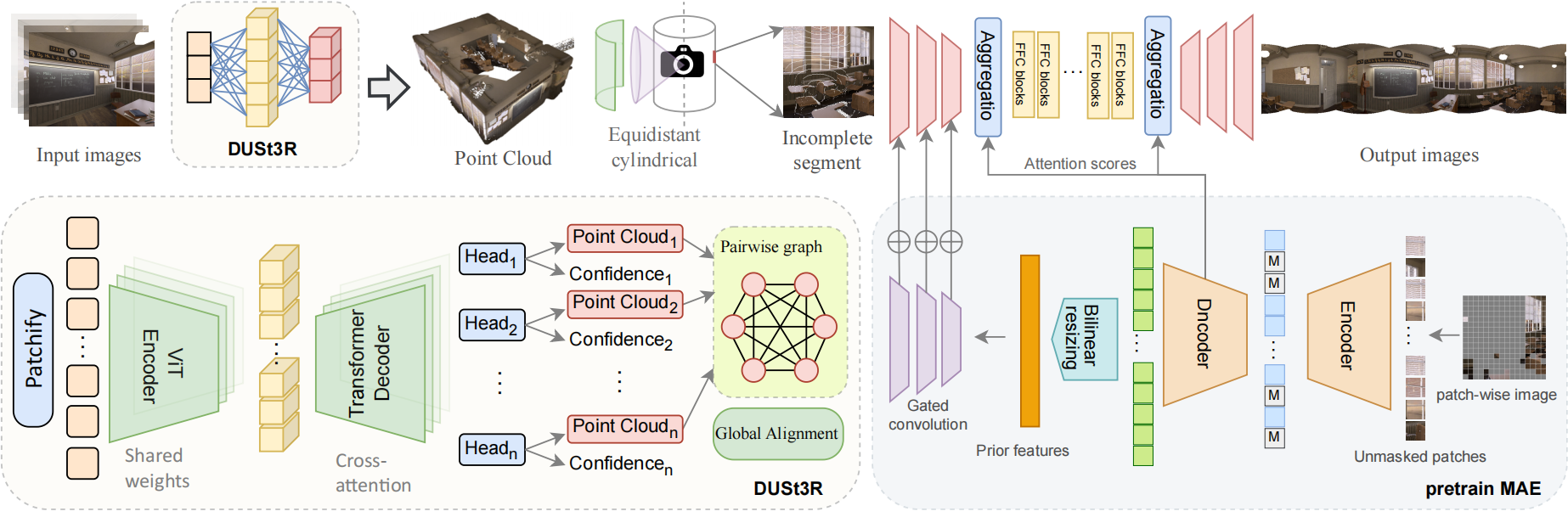}
    \caption{Schematic diagram of the algorithm workflow presented in this paper. The 2D input images are first processed by a reconstruction network to generate corresponding 3D point cloud scenes. For multiple input images, the generated scenes are globally aligned. Then, an equidistant cylindrical projection is applied to reduce the dimensionality of the 3D point clouds onto a 2D manifold. Finally, hole filling is performed on the resulting images. In this paper, based on an improved MAE algorithm, the output of MAE is used as a prior input to the completion network, enabling large-scale semantic completion.}
    \label{fig:flow}
\end{figure*}

Figure \ref{fig:flow} presents a schematic overview of the algorithmic framework proposed in this study. The process begins with a set of $N$ unordered input images, denoted as $\{\mathbf{I}_{i}\}_{i=1}^{N}$, where each image $\mathbf{I}_{i} \in \mathbb{R}^{H \times W \times 3}$. Each image is initially transformed into a dense three-dimensional (3D) point cloud within a common coordinate system via a 3D lifting module, which concurrently produces pixel-level confidence scores to assess the reliability of the geometric estimations. In the current implementation, the DUSt3R method \cite{2024dust3r} is employed for 3D lifting; however, the overall framework is agnostic to the specific choice of this component and can be substituted with alternative approaches such as VGGT \cite{wang2025vggt} without affecting the subsequent stitching methodology.

Following the generation of the fused 3D point set, a unified projection center is established in 3D space. The point cloud is then projected onto a two-dimensional (2D) panoramic manifold through an equidistant cylindrical projection. This approach contrasts with conventional pairwise 2D view deformation techniques by utilizing a single projection center to consistently map all viewpoints, thereby mitigating distortion propagation commonly induced by multi-view accumulation and inherently supporting the creation of $360^{\circ}$ closed-loop panoramic images. Due to the projection of occluded or distant regions—unobservable in the original images—onto the panoramic canvas, the resulting projection typically contains missing areas or holes. To address this, a hole completion module operating in the canvas domain is introduced, which restores these absent regions by integrating global semantic information with local texture details, ultimately producing continuous and geometrically coherent panoramic surround views.

For the sake of clarity and to maintain consistent notation throughout Section 3, the following conventions are adopted. The pixel domain of each input image, with resolution $H \times W$, is defined as
\[
\Omega = \{(u,v) \mid u \in [0, W),\ v \in [0, H) \}.
\]
Image pixel coordinates are represented as $\mathbf{p} = (u,v) \in \Omega$, while 3D point coordinates within the unified coordinate system (also referred to as the world or unified frame) are denoted by $\mathbf{X} \in \mathbb{R}^3$. The 3D lifting module outputs, for each image, a dense point map in the camera coordinate system, $\mathbf{X}^c_i \in \mathbb{R}^{H \times W \times 3}$, along with corresponding pixel-level confidence scores $\mathbf{C}_i \in [0,1]^{H \times W}$. Additionally, rigid body transformations from each camera view to the unified coordinate system, expressed as $T_i = (\mathbf{R}_i, \mathbf{t}_i) \in SE(3)$, are either estimated or externally provided. These transformations enable the conversion of the point maps into the unified coordinate system, yielding $\mathbf{P}_i$ (refer to Eq. \ref{eq:cam_to_world}). By associating each point in $\mathbf{P}_i$ with color and confidence weights, a globally weighted colored point set $\mathcal{Q}$ is constructed (see Eq. \ref{eq:colored_point_set}). The panoramic canvas onto which the points are projected is defined with resolution $W_f \times H_f$ and pixel domain $\Omega_f = [0, W_f) \times [0, H_f)$.

During the dimensionality reduction phase, all 3D points are parameterized directionally relative to a unique unified projection center $O$ (Eq. \ref{eq:optical_center}). An equidistant cylindrical projection function, denoted $\pi_{\mathrm{cyl}}(\cdot)$, maps these 3D points onto 2D canvas coordinates $\mathbf{q} = [x, y]^{\top} \in \Omega_f$ (Eq. \ref{eq:pi_cyl_def}). Subsequently, a 2D kernel function $K(\cdot)$ is employed to perform normalized weighted accumulation over the discrete set of projected points (Eq. \ref{eq:splatting}), resulting in the initial panoramic canvas $\mathbf{Y}: \Omega_f \rightarrow \mathbb{R}^3$. The accumulation weights $Z(\mathbf{q})$ define a hole mask $\mathbf{M}: \Omega_f \rightarrow \{0,1\}$ (Eq. \ref{eq:mask_def}), which serves as a clear and consistent input for the subsequent hole completion module operating within the canvas domain.

\subsection{Analysis of Distortion in Two-Dimensional Spatial Mapping}
\label{sec:distortion-in-2d}

Current image stitching frameworks encounter two intrinsic and theoretically unavoidable challenges: projection distortion induced by the homography matrix and ghosting artifacts arising in scenes with significant parallax. This section first examines the projection distortion issue attributable to the homography matrix.

In conventional image stitching methodologies, the homography matrix serves as a fundamental component for image projection transformations. Formally, a homography matrix is a non-singular \(3 \times 3\) matrix that characterizes a projective mapping between two planes. It facilitates the correspondence of points from one image to another, thereby enabling pixel alignment and stitching. However, the theoretical validity of homography transformations is confined to two specific conditions: either the corresponding points reside on the same planar surface, or the camera centers coincide during image acquisition \cite{2007Automatic}. In practical scenarios where these conditions are violated, employing a homography matrix to project and stitch pixels from two images inevitably results in distortion.

Consider two input images \(I\) and \(I'\), where a point \((x,y)\) in the source image \(I\) corresponds to a point \((x', y')\) in the target image \(I'\). Given a homography matrix \(\hat{H}\), the mapping from \((x,y)\) in the original image coordinate system to \((x', y')\) in the target image coordinate system can be expressed in homogeneous coordinates as:
\[
\begin{bmatrix} x' \\ y' \\ 1 \end{bmatrix} 
\sim 
\begin{bmatrix} 
\hat{h}_1 & \hat{h}_2 & \hat{h}_3 \\ 
\hat{h}_4 & \hat{h}_5 & \hat{h}_6 \\ 
\hat{h}_7 & \hat{h}_8 & 1 
\end{bmatrix}
\begin{bmatrix} x \\ y \\ 1 \end{bmatrix}.
\]
This relationship can equivalently be written as:
\[
x' = \frac{\hat{h}_1 x + \hat{h}_2 y + \hat{h}_3}{\hat{h}_7 x + \hat{h}_8 y + 1}, \quad 
y' = \frac{\hat{h}_4 x + \hat{h}_5 y + \hat{h}_6}{\hat{h}_7 x + \hat{h}_8 y + 1}.
\]

To gain a more intuitive understanding of the relationship between the projection transformation and pixel coordinates, we adopt the approach of Chum et al. \cite{2005The} by rotating the coordinate system \((x,y)\) by an angle \(\beta\) to obtain new coordinates \((u,v)\), defined as:
\[
\begin{bmatrix} x \\ y \end{bmatrix} = 
\begin{bmatrix} \cos \beta & -\sin \beta \\ \sin \beta & \cos \beta \end{bmatrix}
\begin{bmatrix} u \\ v \end{bmatrix},
\]
where
\[
\beta = \mathrm{atan2}(-\hat{h}_8, -\hat{h}_7).
\]
This rotation effectively nullifies \(\hat{h}_8\) in the homography matrix, yielding a transformed homography matrix:
\[
\begin{bmatrix} x' \\ y' \\ 1 \end{bmatrix} 
\sim 
\begin{bmatrix}
h_1 & h_2 & h_3 \\
h_4 & h_5 & h_6 \\
-c & 0 & 1
\end{bmatrix}
\begin{bmatrix} u \\ v \\ 1 \end{bmatrix},
\]
where the parameters satisfy:
\[
c = \sqrt{\hat{h}_7^2 + \hat{h}_8^2},
\]
\[
\begin{bmatrix} h_1 & h_2 \\ h_4 & h_5 \end{bmatrix} = 
\begin{bmatrix} \hat{h}_1 & \hat{h}_2 \\ \hat{h}_4 & \hat{h}_5 \end{bmatrix}
\begin{bmatrix} \cos \beta & -\sin \beta \\ \sin \beta & \cos \beta \end{bmatrix},
\]
and
\[
(h_3, h_6) = (\hat{h}_3, \hat{h}_6).
\]

Within this rotated \((u,v)\) coordinate system, the coordinate mapping can be reformulated as:
\[
x' = \frac{h_1 u + h_2 v + h_3}{1 - c u}, \quad 
y' = \frac{h_4 u + h_5 v + h_6}{1 - c u}.
\]
Notably, the denominator depends solely on the coordinate \(u\), enabling focused analysis of the homography transformation’s dependence on \(u\).

The homography matrix \(H\) can be decomposed into the product of an affine transformation \(H_A\) and a pure projective transformation \(H_P\):
\begin{equation}
    \begin{split}
        H &= 
\begin{bmatrix}
h_1 & h_2 & h_3 \\
h_4 & h_5 & h_6 \\
-c & 0 & 1
\end{bmatrix}\\ &= 
\begin{bmatrix}
h_1 + c h_3 & h_2 & h_3 \\
h_4 + c h_6 & h_5 & h_6 \\
0 & 0 & 1
\end{bmatrix}
\begin{bmatrix}
1 & 0 & 0 \\
0 & 1 & 0 \\
-c & 0 & 1
\end{bmatrix}\\ &= H_A H_P.
    \end{split}
\end{equation}

The determinant of the Jacobian matrix characterizes the local area scaling induced by the transformation. Analyzing the Jacobian determinant of \(H\) yields:
\[
\det J(u,v) = \det J_A(u,v) \cdot \frac{1}{(1 - c u)^3},
\]
where the affine component’s Jacobian determinant is a constant:
\[
s_A = (h_1 + c h_3) h_5 - (h_4 + c h_6) h_2,
\]
which is independent of \(u\) and \(v\). This indicates that the affine transformation scales areas uniformly across the coordinate space. In contrast, the projective component introduces a denominator dependent on \(u\), causing the area scaling distortion to intensify as \(u\) increases.

\begin{figure}[htbp]
\centering
\subfloat[SIFT+RANSAC]{
		\includegraphics[scale=0.21]{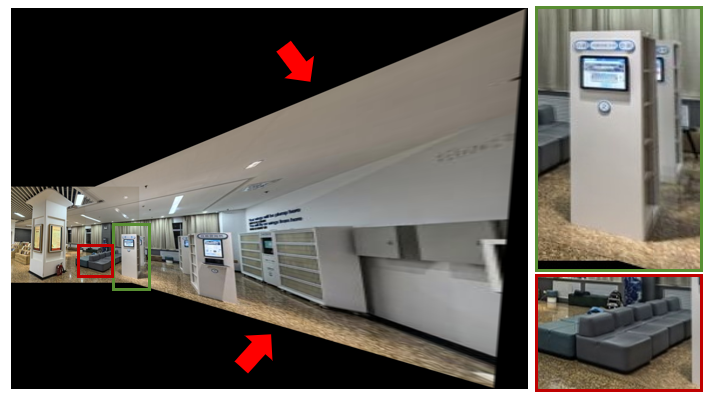}}\vspace{3pt}
\subfloat[UDIS++]{
		\includegraphics[scale=0.21]{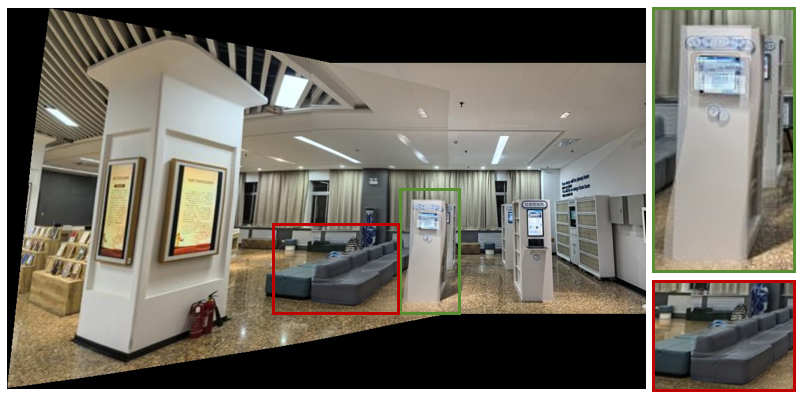}}
\caption{Illustration of projection distortion. (a) Stitching result using Autostitch: the overlapping region exhibits minimal ghosting artifacts, whereas the non-overlapping region suffers from pronounced stretching distortion, as indicated by the red arrows. (b) Stitching result using UDIS++ \cite{2023udis++}: distortion in the non-overlapping region is reduced, albeit at the expense of conspicuous ghosting artifacts within the overlapping region.}
\label{fig:projection-distortion}
\end{figure}

This theoretical insight is corroborated by empirical observations, as illustrated in Figure \ref{fig:projection-distortion}. When employing basic stitching methods such as SIFT combined with RANSAC, the projected image exhibits progressively severe stretching distortions away from the overlapping regions.

The preceding analysis assumes \(h_8 = 0\), thereby eliminating the influence of the \(v\) coordinate and isolating the homography’s effect along the \(u\) axis. Analogously, setting \(h_7 = 0\) allows examination of distortion along the \(v\) axis. Consequently, it can be concluded that the homography matrix inherently induces stretching distortions along two orthogonal directions.

In the domain of image stitching, inspired by Autostitch \cite{2007Automatic}, the majority of existing algorithms—both traditional and deep learning-based—rely on homography matrices for image alignment. Although homography-based methods perform adequately on certain datasets for two-dimensional image alignment, their intrinsic theoretical limitations ensure that projection distortion remains an unavoidable issue. Approaches such as SPHP \cite{2014Sphp}, AANAP \cite{2015Aanap}, and GSP \cite{2016gsp} can only alleviate but not eliminate projection distortion. Given the relative scarcity of comprehensive image stitching datasets, current algorithms primarily employ heuristic remedies to mitigate these theoretical shortcomings; nonetheless, distortion persists when real-world scenes deviate from the idealized assumptions.

Another critical challenge arises from ghosting artifacts in scenes characterized by large parallax, which homography-based mappings cannot adequately address. The homography matrix presupposes that all scene elements lie on a single plane; however, real-world scenes typically contain objects at varying depths. Depth discontinuities in three-dimensional space introduce significant singularities. Most existing algorithms attempt to identify such regions and apply gradual fitting strategies to circumvent singularities, as exemplified by the UDIS++ \cite{2023udis++} stitching results depicted in Figure \ref{fig:projection-distortion}. While these methods can partially constrain projection distortion, the necessity for pixels in overlapping regions to satisfy constraints from multiple viewpoints inherently leads to ghosting artifacts when images are composited onto a common canvas.

In scenes with substantial parallax, the camera’s proximity to objects accentuates the influence of depth variations on stitching quality and accuracy. This effect is less pronounced in two-image stitching, where distortion in non-overlapping regions of the secondary view can be sacrificed to preserve overlapping region quality. However, in multi-image sequential stitching, intermediate views must simultaneously satisfy constraints from adjacent images, resulting in irreconcilable distortions. Projection errors stemming from depth inconsistencies accumulate, exacerbating ghosting artifacts and potentially causing structural collapse of the panoramic image, thereby failing to faithfully represent the real scene. These ghosting artifacts significantly degrade stitching quality and reliability, imposing limitations on traditional two-dimensional image stitching methods in complex environments.

Fundamentally, the aforementioned stitching challenges stem from the insufficiency of two-dimensional representations. Consequently, issues manifested in the panorama cannot be fully resolved within a purely two-dimensional spatial framework. From a theoretical standpoint, a comprehensive solution necessitates expressing each image’s two-dimensional pixels through inverse transformations involving pixel coordinate systems, normalized image plane coordinates, camera coordinate systems, and world coordinate systems—thereby unifying all images within a common world coordinate system—followed by forward panoramic projection. Such an approach is essential to fundamentally overcome projection distortion problems.

\subsection{Transformation of 2D Pixel Representations into 3D Point Cloud Representations}
\label{sec:lift-2d-to-3d}

As outlined in Section \ref{sec:distortion-in-2d}, a fundamental limitation of two-dimensional deformation techniques arises from the fact that both their optimization variables and constraints are confined to the 2D image plane. This restriction impedes the explicit enforcement of geometric consistency across varying depth layers. Consequently, under conditions of pronounced parallax and occlusion, artifacts such as structural bending, ghosting, and localized mismatches frequently manifest. To address these challenges and establish robust cross-view three-dimensional correspondences, the present study adopts a “lift first, then stitch, and finally reproject” methodology. Specifically, each input image is initially transformed into a dense 3D point representation, subsequently aligned within a unified three-dimensional coordinate system, and ultimately projected onto the target two-dimensional manifold to produce a panoramic composite.

\paragraph{3D Lifting Operator and Output Specification}

We conceptualize the 3D lifting process as an operator denoted by $\mathcal{G}$. Given the $i$-th input image $\mathbf{I}_{i} \in \mathbb{R}^{H \times W \times 3}$, the operator $\mathcal{G}$ generates a dense point map expressed in the camera coordinate frame, accompanied by a pixel-wise confidence map:
\begin{equation}
(\mathbf{X}^{c}_{i},\ \mathbf{C}_{i}) = \mathcal{G}(\mathbf{I}_{i}),
\quad
\mathbf{X}^{c}_{i} \in \mathbb{R}^{H \times W \times 3},\quad \mathbf{C}_{i} \in [0,1]^{H \times W}.
\label{eq:lifting_operator}
\end{equation}
Here, $\mathbf{X}^{c}_{i}(u,v) = [x^{c}_{i}(u,v),\, y^{c}_{i}(u,v),\, z^{c}_{i}(u,v)]^{\top} \in \mathbb{R}^3$ represents the three-dimensional spatial coordinates in the camera frame corresponding to the pixel location $\mathbf{p} = (u,v)$. The scalar $\mathbf{C}_{i}(u,v)$ denotes a confidence weight associated with this point, taking continuous values within the interval $[0,1]$, where higher values signify greater reliability of the geometric estimation. It is critical to note that $\mathbf{C}_{i}$ functions as a soft weighting factor rather than a binary mask; this facilitates subsequent filtering and weighting operations aimed at mitigating the influence of weakly textured regions, dynamic objects, or occlusion boundaries during the fusion process.

\paragraph{Alignment within a Unified Coordinate System}

To enable global three-dimensional stitching, it is necessary to transform the point maps from each viewpoint into a common coordinate system. Let $T_i = (\mathbf{R}_i, \mathbf{t}_i) \in SE(3)$ denote the rigid body transformation mapping points from the camera coordinate system of the $i$-th image to the unified coordinate frame. Accordingly, the 3D point corresponding to pixel $(u,v)$ satisfies
\begin{equation}
\mathbf{P}_{i}(u,v) = \mathbf{R}_{i} \mathbf{X}^{c}_{i}(u,v) + \mathbf{t}_{i},
\label{eq:cam_to_world}
\end{equation}
where $\mathbf{P}_{i}(u,v) \in \mathbb{R}^3$ denotes the transformed point in the unified coordinate system. Our framework requires only that the operator $\mathcal{G}$ provide dense 3D points and associated confidence values, as well as the capability to estimate or externally obtain the set of multi-view alignment transformations $\{T_i\}$. In our implementation, we instantiate $\mathcal{G}$ using DUSt3R \cite{2024dust3r}, leveraging its multi-view correspondences to solve for $\{T_i\}$. Nonetheless, the proposed approach is agnostic to the specific 3D lifting method employed; alternative models such as VGGT \cite{wang2025vggt} may be substituted without altering the fundamental spatial lifting and stitching paradigm described herein.

\paragraph{Confidence-Based Filtering and Fusion Weighting}

The generated 3D point maps inherently contain noise and regions of local instability. To address this, we first apply a confidence-based filtering criterion, defining the set of valid pixels as
\begin{equation}
\Omega_i = \left\{ (u,v) \in \Omega \ \big|\ \mathbf{C}_{i}(u,v) \ge \tau_c \right\},
\label{eq:valid_set}
\end{equation}
where $\tau_c$ is a predefined threshold. Additionally, to suppress contributions from points near depth discontinuities and occlusion boundaries, we introduce a local geometric variation metric:
\begin{equation}
s_i(u,v) = \left\| \nabla \mathbf{P}_{i}(u,v) \right\|_F,
\quad
\nabla \mathbf{P}_{i}(u,v) = \left[ \frac{\partial \mathbf{P}_{i}}{\partial u},\ \frac{\partial \mathbf{P}_{i}}{\partial v} \right],
\label{eq:local_geom_var}
\end{equation}
where $\|\cdot\|_F$ denotes the Frobenius norm, and $\nabla \mathbf{P}_{i}(u,v)$ represents the first-order spatial derivatives of the point map with respect to pixel coordinates, approximated in practice via finite differences. This measure attains low values in regions of continuous depth but increases markedly near occlusion boundaries or geometric discontinuities. Based on this metric, fusion weights for each valid point are assigned as
\begin{equation}
w_i(u,v) = \mathbf{C}_{i}(u,v) \cdot \rho\bigl(s_i(u,v)\bigr),
\label{eq:fusion_weight}
\end{equation}
where $\rho(\cdot)$ is a monotonically decreasing robust function—examples include $\rho(s) = \exp(-s/\sigma)$ or $\rho(s) = 1/(1+s)$—designed to attenuate the influence of geometrically unstable regions during fusion.

\paragraph{Formation of a Weighted Colored Point Set}

By integrating the 3D point coordinates, corresponding color information, and fusion weights into a unified data structure, subsequent operations such as dimensionality reduction, manifold projection, and canvas domain completion can be consistently and coherently managed at the input stage. Concretely, for each valid pixel $(u,v) \in \Omega_i$, the color vector $\mathbf{I}_{i}(u,v) \in \mathbb{R}^3$ is associated with its corresponding 3D point $\mathbf{P}_{i}(u,v)$, yielding the global weighted colored point set
\begin{equation}
\mathcal{Q} = \bigcup_{i=1}^N \left\{ \left( \mathbf{P}_{i}(u,v),\ \mathbf{I}_{i}(u,v),\ w_i(u,v) \right) \ \big|\ (u,v) \in \Omega_i \right\}.
\label{eq:colored_point_set}
\end{equation}
This composite point set $\mathcal{Q}$ serves as the unified input for the two-dimensional manifold projection detailed in Section \ref{sec:3d-to-2d-manifold} and the subsequent canvas generation and hole completion procedures described in Section \ref{sec:hole-completion}, thereby establishing a clear and consistent data flow throughout the entire algorithmic pipeline.

\subsection{Dimensionality Reduction of Three-Dimensional Point Clouds onto a Two-Dimensional Manifold}
\label{sec:3d-to-2d-manifold}

In Section \ref{sec:lift-2d-to-3d}, we derived a weighted, color-annotated point set \(\mathcal{Q}\) expressed within a unified coordinate framework. The objective of the present section is to project \(\mathcal{Q}\) onto a two-dimensional panoramic canvas domain \(\Omega_f\) via a single, unified projection center, and to rigorously define the associated projection operator \(\pi_{\mathrm{cyl}}(\cdot)\). Distinct from conventional pairwise two-dimensional deformation stitching methods, our approach parameterizes the directions of all three-dimensional points with respect to a common projection center. This strategy circumvents loop discontinuities that typically arise from projection center drift in multi-view stitching scenarios.

\paragraph{Unified Projection Center}

Let \(\mathbf{t}_i\) denote the camera center of the \(i\)-th view within the unified coordinate system (refer to Equation \ref{eq:cam_to_world}, where \(\mathbf{t}_i\) represents the origin of the camera coordinate system after transformation). We define the unified projection center \(O\) as the arithmetic mean of all camera centers:

\begin{equation}
O \;=\; \frac{1}{N}\sum_{i=1}^{N}\mathbf{t}_i \ \in \mathbb{R}^{3}.
\label{eq:optical_center}
\end{equation}

This definition exhibits stability in multi-view circular capture configurations: as the cameras traverse a circular path around the scene, the point \(O\) approximately resides within the enclosed trajectory, thereby serving as a consistent virtual optical center.

\paragraph{Angular Parameterization via Equidistant Cylindrical Projection}

For an arbitrary three-dimensional point \(\mathbf{X} \in \mathbb{R}^3\), we define its direction vector relative to \(O\) as

\begin{equation}
\mathbf{d}(\mathbf{X}) \;=\; \mathbf{X}-O \;=\; [d_x,\ d_y,\ d_z]^{\top}.
\label{eq:dir_vec}
\end{equation}

This direction is parameterized by the azimuth angle \(\theta\) and the elevation angle \(\phi\), given by

\begin{equation}
\begin{split}
\theta(\mathbf{X})&=\operatorname{atan2}(d_y,\ d_x)\in(-\pi,\pi],\\
\phi(\mathbf{X})&=\operatorname{atan2}\!\left(d_z,\ \sqrt{d_x^2+d_y^2}\right)\in\left(-\tfrac{\pi}{2},\tfrac{\pi}{2}\right).
\label{eq:angles}
\end{split}
\end{equation}

Here, \(\theta\) represents the horizontal azimuth angle with its inherent \(2\pi\)-periodicity, while \(\phi\) corresponds to the vertical elevation angle.

\paragraph{Two-Dimensional Manifold Interpretation: Cylindrical Surface Embedding and Parameterization}

As previously noted, the panoramic canvas can be interpreted as a parameterization of a two-dimensional manifold embedded within three-dimensional Euclidean space. We provide a geometric characterization of this cylindrical manifold. Taking \(O\) as a reference point along the cylinder’s central axis and fixing a radius \(R > 0\), the cylindrical surface is defined by

\begin{equation}
\mathcal{M}_{\mathrm{cyl}} = \left\{ \mathbf{s} \in \mathbb{R}^3 \mid (s_x - O_x)^2 + (s_y - O_y)^2 = R^2 \right\},
\label{eq:cyl_manifold}
\end{equation}

which constitutes a two-dimensional manifold in \(\mathbb{R}^3\). For any point \(\mathbf{X}\), the intersection of the ray emanating from \(O\) in the direction \(\mathbf{d}(\mathbf{X})\) with the cylindrical surface is expressed as

\begin{equation}
\mathbf{s}(\mathbf{X}) = O + \alpha (\mathbf{X} - O), \quad \text{where} \quad \alpha = \frac{R}{\sqrt{d_x^2 + d_y^2} + \epsilon},
\label{eq:cyl_intersection}
\end{equation}

with \(\epsilon\) being a small positive constant introduced to ensure numerical stability. The local coordinates of this intersection on the cylinder are naturally given by the azimuth angle \(\theta\) (as defined above) and the axial height

\[
h = s_z(\mathbf{X}) - O_z = \alpha d_z = R \tan \phi(\mathbf{X}).
\]

Consequently, the pair \((\theta, \phi)\) serves as an equivalent parameterization of the cylindrical manifold, with the vertical coordinate \(h\) related to \(\phi\) via a monotonic transformation. In this study, we employ an equidistant encoding scheme that utilizes \(\phi\) as the vertical parameter. This approach uniformly samples the vertical angular domain, thereby yielding more stable panoramic layouts and consistent visual scales, particularly under wide field-of-view conditions.

\paragraph{Mapping from Angular Coordinates to Canvas Coordinates}

Let the panoramic canvas have a resolution of \(W_f \times H_f\), with pixel domain \(\Omega_f = [0, W_f) \times [0, H_f)\). The equidistant cylindrical projection maps the angular coordinates \((\theta, \phi)\) linearly onto canvas coordinates \(\mathbf{q} = [x, y]^\top\) as follows:

\begin{equation}
x = \frac{W_f}{2\pi} \bigl(\theta(\mathbf{X}) + \pi \bigr), \quad
y = \frac{H_f}{\pi} \left( \frac{\pi}{2} - \phi(\mathbf{X}) \right).
\label{eq:cyl_uv}
\end{equation}

Here, the coordinate \(x\) corresponds to the horizontal azimuthal direction, while \(y\) corresponds to the vertical direction. Since \(\theta + \pi \in [0, 2\pi)\) spans the full \(360^\circ\) field of view, modular arithmetic with modulus \(W_f\) can be applied near the horizontal boundaries to accommodate periodicity and ensure seamless loop closure.

\paragraph{Formal Definition of the Projection Operator}

In summary, the equidistant cylindrical projection is formalized as the mapping

\begin{equation}
\pi_{\mathrm{cyl}}: \mathbb{R}^3 \to \Omega_f, \quad \pi_{\mathrm{cyl}}(\mathbf{X}) = \begin{bmatrix} x \\ y \end{bmatrix},
\label{eq:pi_cyl_def}
\end{equation}

where the coordinates \((x, y)\) are computed according to the relations specified in Equations \ref{eq:angles} through \ref{eq:cyl_uv}. Subsequent processes, including canvas generation (cf. Equation \ref{eq:splatting}) and hole detection (cf. Equation \ref{eq:mask_def}), utilize \(\pi_{\mathrm{cyl}}(\mathbf{X})\) as the target location of each three-dimensional point on the panoramic canvas. Kernel-weighted accumulation techniques are employed to robustly render the discrete point set into a continuous image representation.

\subsection{Hole Completion in Two-Dimensional Manifold Projection}
\label{sec:hole-completion}

\begin{figure}[htbp]
    \centering
    \includegraphics[width=0.65\linewidth]{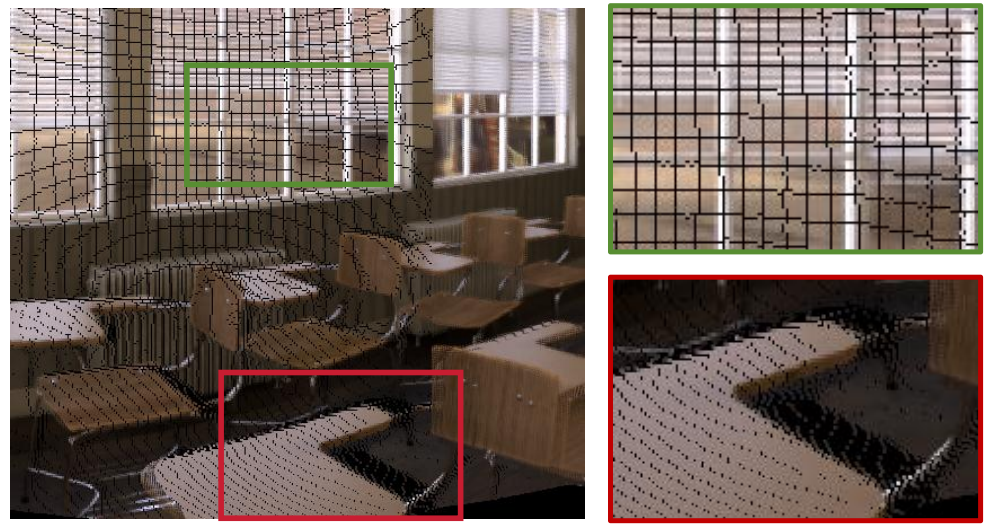}
    \caption{Visualization of projection-induced holes. The green box highlights regular holes attributable to discrete sampling, whereas the red box emphasizes larger holes resulting from changes in viewpoint and occlusion.}
    \label{fig:hole}
\end{figure}

Utilizing the isometric cylindrical projection function $\pi_{\mathrm{cyl}}(\cdot)$ introduced in Section~\ref{sec:3d-to-2d-manifold}, we project the weighted, colored point set $\mathcal{Q}$ onto a panoramic canvas domain to generate an initial panoramic image alongside its corresponding visibility mask. Due to constraints such as limited input viewpoints, alterations in occlusion relationships, and incomplete coverage arising from discrete sampling, inevitable holes manifest in the projection, as depicted in Figure~\ref{fig:hole}. These voids not only compromise the visual completeness but also disrupt the structural coherence of the panorama. Consequently, it is imperative to perform completion directly within the canvas domain rather than relying solely on geometric interpolation techniques.

\paragraph{Canvas Construction and Definition of Hole Mask}

Following the completion of geometric alignment and fusion in three-dimensional space, it is necessary to robustly render the colored 3D point set onto a two-dimensional panoramic canvas. To achieve this, we employ a normalized kernel-weighted projection accumulation approach, which can be interpreted as a form of splatting rendering. Each 3D point $\mathbf{X} \in \mathbb{R}^3$ is first mapped to a continuous 2D coordinate $\pi_{\mathrm{cyl}}(\mathbf{X}) \in \mathbb{R}^2$ on the canvas via the isometric cylindrical projection $\pi_{\mathrm{cyl}}(\cdot)$. Subsequently, a two-dimensional kernel function $K(\cdot)$ diffuses the contribution of each point to its neighborhood in the projection domain, thereby mitigating aliasing effects and reducing holes caused by discrete sampling.

Let $\mathcal{Q}$ denote the set of points with associated attributes used for rendering, where each element is a triplet $(\mathbf{X}, \mathbf{c}, w)$: $\mathbf{X}$ represents the 3D position in the unified coordinate system, $\mathbf{c} \in \mathbb{R}^3$ denotes the color or texture vector, and $w \in \mathbb{R}_{\geq 0}$ is a fusion weight encoding the point’s reliability, visibility, or reprojection consistency. For any canvas coordinate $\mathbf{q} = [x, y]^\top$, the pixel value $\mathbf{Y}(\mathbf{q})$ on the canvas is computed as follows:

\begin{equation}
\mathbf{Y}(\mathbf{q})
=
\frac{
\sum\limits_{(\mathbf{X}, \mathbf{c}, w) \in \mathcal{Q}}
w \, K\!\left(\mathbf{q} - \pi_{\mathrm{cyl}}(\mathbf{X})\right) \, \mathbf{c}
}{
\sum\limits_{(\mathbf{X}, \mathbf{c}, w) \in \mathcal{Q}}
w \, K\!\left(\mathbf{q} - \pi_{\mathrm{cyl}}(\mathbf{X})\right)
+ \varepsilon
}
\label{eq:splatting}
\end{equation}

Here, the vector $\mathbf{q} - \pi_{\mathrm{cyl}}(\mathbf{X}) \in \mathbb{R}^2$ represents the offset between the canvas coordinate and the projected position of the 3D point. The kernel function $K(\cdot)$ yields a non-negative scalar weight that decreases as the offset increases, thereby controlling the spatial influence of each 3D point on neighboring pixels. The numerator aggregates the weighted colors of all points contributing to position $\mathbf{q}$, with the fusion weights $w$ attenuating the impact of unreliable points. The denominator normalizes these weights to produce a weighted average, which prevents brightness or color bias due to local variations in point density. The small positive constant $\varepsilon$ ensures numerical stability in cases where no points contribute to a given pixel.

Based on this accumulation process, we define a hole mask to identify unobservable regions in the projection. Specifically, the accumulated observation support strength at canvas coordinate $\mathbf{q} \in \Omega_f$ is defined as

\begin{equation}
Z(\mathbf{q})
=
\sum\limits_{(\mathbf{X}, \mathbf{c}, w) \in \mathcal{Q}}
w \, K\!\left(\mathbf{q} - \pi_{\mathrm{cyl}}(\mathbf{X})\right),
\label{eq:accum_weight}
\end{equation}

where $K(\cdot)$ is the kernel function and $\pi_{\mathrm{cyl}}(\mathbf{X})$ denotes the projected location of the 3D point on the canvas. A higher value of $Z(\mathbf{q})$ indicates a greater concentration of reliable observations from the original multi-view data near that position. Conversely, when changes in viewpoint or occlusion removal cause previously occluded distant regions to be projected onto the canvas without corresponding original observations, $Z(\mathbf{q})$ becomes markedly lower. Accordingly, we define a binary hole mask with threshold $\tau > 0$ as

\begin{equation}
\mathbf{M}(\mathbf{q}) =
\begin{cases}
1, & Z(\mathbf{q}) < \tau, \\
0, & Z(\mathbf{q}) \geq \tau,
\end{cases}
\label{eq:mask_def}
\end{equation}

where $\mathbf{M}(\mathbf{q}) = 1$ designates a hole region requiring completion in subsequent processing, and $\mathbf{M}(\mathbf{q}) = 0$ indicates a position with sufficient observational support that should remain unaltered during completion.

\paragraph{Completion Model in the Canvas Domain}

Let the initial panoramic canvas be denoted by the function $\mathbf{Y} : \Omega_f \rightarrow \mathbb{R}^3$, where $\Omega_f$ represents the pixel domain of the canvas and $\mathbf{Y}(\mathbf{q})$ is the RGB color vector at coordinate $\mathbf{q} = [x, y]^\top$. The corresponding hole mask is denoted by $\mathbf{M} : \Omega_f \rightarrow \{0,1\}$, with $\mathbf{M}(\mathbf{q}) = 1$ indicating regions lacking effective observational support and thus requiring completion, and $\mathbf{M}(\mathbf{q}) = 0$ indicating reliably observed regions to be preserved. For convenience, define the complement mask $\bar{\mathbf{M}} = \mathbf{1} - \mathbf{M}$, where $\mathbf{1}$ is the all-ones mask, and let $\odot$ denote element-wise multiplication broadcast across the three color channels.

We introduce a completion operator $\mathcal{F}_{\theta}$ acting within the canvas domain, designed to infer plausible textures within hole regions while maintaining consistency in observed areas. The input to this operator is constructed by retaining the observed regions and zeroing out the hole regions:

\begin{equation}
\label{eq:inpaint_input}
\mathbf{Y}^{\mathrm{in}} = \bar{\mathbf{M}} \odot \mathbf{Y},
\end{equation}

which, together with the hole mask $\mathbf{M}$, is fed into the completion model to produce the predicted canvas:

\begin{equation}
\label{eq:inpaint_operator}
\hat{\mathbf{Y}} = \mathcal{F}_{\theta}\!\left(\mathbf{Y}^{\mathrm{in}}, \mathbf{M}\right).
\end{equation}

In this formulation, $\hat{\mathbf{Y}}(\mathbf{q})$ provides the completed texture for pixels where $\mathbf{M}(\mathbf{q}) = 1$, while for pixels with $\mathbf{M}(\mathbf{q}) = 0$, the output should ideally remain consistent with the original $\mathbf{Y}(\mathbf{q})$. The final output during inference is obtained by fusing the observed and completed regions as follows:

\begin{equation}
\label{eq:inpaint_fuse}
\mathbf{Y}^{\star} = \bar{\mathbf{M}} \odot \mathbf{Y} + \mathbf{M} \odot \hat{\mathbf{Y}}.
\end{equation}

This fusion explicitly ensures that observed regions are preserved from the 3D projection accumulation results, while completion is confined to hole regions, thereby preventing unnecessary alterations in already observed textures.

Given the scarcity of complete, hole-free panoramic ground truth in real-world data, we adopt a self-supervised masked reconstruction strategy to train $\mathcal{F}_{\theta}$ within the canvas domain. Specifically, on the existing canvas $\mathbf{Y}$, we sample an additional random occlusion mask $\mathbf{R} : \Omega \rightarrow \{0,1\}$, where $\mathbf{R}(\mathbf{q}) = 1$ indicates artificially occluded pixels. We then construct a combined mask:

\begin{equation}
\label{eq:joint_mask}
\mathbf{M}' = \mathbf{M} \lor \mathbf{R},
\end{equation}

where $\lor$ denotes the pixel-wise logical OR operation. The corresponding network input is

\begin{equation}
\label{eq:masked_input_for_ssl}
\mathbf{Y}^{\mathrm{in}}_{\mathrm{ssl}} = (\mathbf{1} - \mathbf{M}') \odot \mathbf{Y},
\quad
\tilde{\mathbf{Y}} = \mathcal{F}_{\theta}\!\left(\mathbf{Y}^{\mathrm{in}}_{\mathrm{ssl}}, \mathbf{M}'\right).
\end{equation}

Since $\mathbf{R}$ masks originally observable regions (i.e., where $\bar{\mathbf{M}}(\mathbf{q}) = 1$), we can impose reconstruction supervision on these artificially occluded pixels via the loss function:

\begin{equation}
\label{eq:reconstruction_loss_ssl}
\mathcal{L}_{\mathrm{rec}} = \left\| \mathbf{R} \odot \bigl(\tilde{\mathbf{Y}} - \mathbf{Y}\bigr) \right\|_1,
\end{equation}

where $\|\cdot\|_1$ denotes the sum of pixel-wise and channel-wise $L_1$ norms. To further prevent color drift in the completion network on genuinely observed regions, we introduce an observation consistency loss that enforces output-input agreement on the original observed region $\bar{\mathbf{M}}$:

\begin{equation}
\label{eq:observed_consistency_loss}
\mathcal{L}_{\mathrm{obs}} = \left\| \bar{\mathbf{M}} \odot \bigl(\tilde{\mathbf{Y}} - \mathbf{Y}\bigr) \right\|_1.
\end{equation}

The overall training objective is thus formulated as

\begin{equation}
\label{eq:final_inpaint_loss}
\mathcal{L} = \mathcal{L}_{\mathrm{rec}} + \lambda \, \mathcal{L}_{\mathrm{obs}},
\end{equation}

where $\lambda$ is a hyperparameter balancing the trade-off between completion accuracy and fidelity to observed data.

The completion model $\mathcal{F}_{\theta}$ functions as a generic module within the canvas domain, with its input-output interface strictly defined by Equations~(\ref{eq:inpaint_operator}) through~(\ref{eq:inpaint_fuse}), independent of the specific network architecture employed. In our implementation, we instantiate $\mathcal{F}_{\theta}$ using a well-established masked autoencoder-based inpainting network. Empirical results demonstrate that this module effectively recovers unobservable regions caused by viewpoint changes, while the principal improvements in overall performance stem from the earlier described framework of 3D-consistent stitching and unified projection.

\section{Experimental Results}
\subsection{Dataset}
The majority of existing panoramic stitching datasets primarily address the stitching of two images. While some datasets encompass multiple images, their fields of view are relatively limited and do not adequately represent real-world application scenarios. To date, there is no comprehensive dataset specifically tailored for scenarios involving substantial parallax. To facilitate a rigorous quantitative evaluation of the algorithm proposed herein, we have developed a bespoke dataset termed MVIS (Multi-View Image Set for Image Stitching)\footnote{The dataset will be made publicly accessible at https://www.kaggle.com/datasets/...}. This dataset comprises two components: synthetically rendered data and real-world captured data. The synthetic portion was generated using Blender software, leveraging the SphereCraft 3D scene library, which enables precise manipulation of camera parameters and scene configurations, as well as provision of accurate ground truth depth and camera pose information. The real-world data were acquired via the rear main camera of a Huawei Mate70 smartphone, encompassing diverse environments such as building corridors, cafes, and libraries. Each panoramic set consists of 7 to 8 images. Notably, the real dataset includes numerous scenes characterized by pronounced parallax and intricate occlusion relationships, thereby serving as a robust testbed for assessing the algorithm’s capacity to manage multiple depth layers and occlusions.

\begin{figure*}[htbp]
    \centering
    \includegraphics[width=0.8\linewidth]{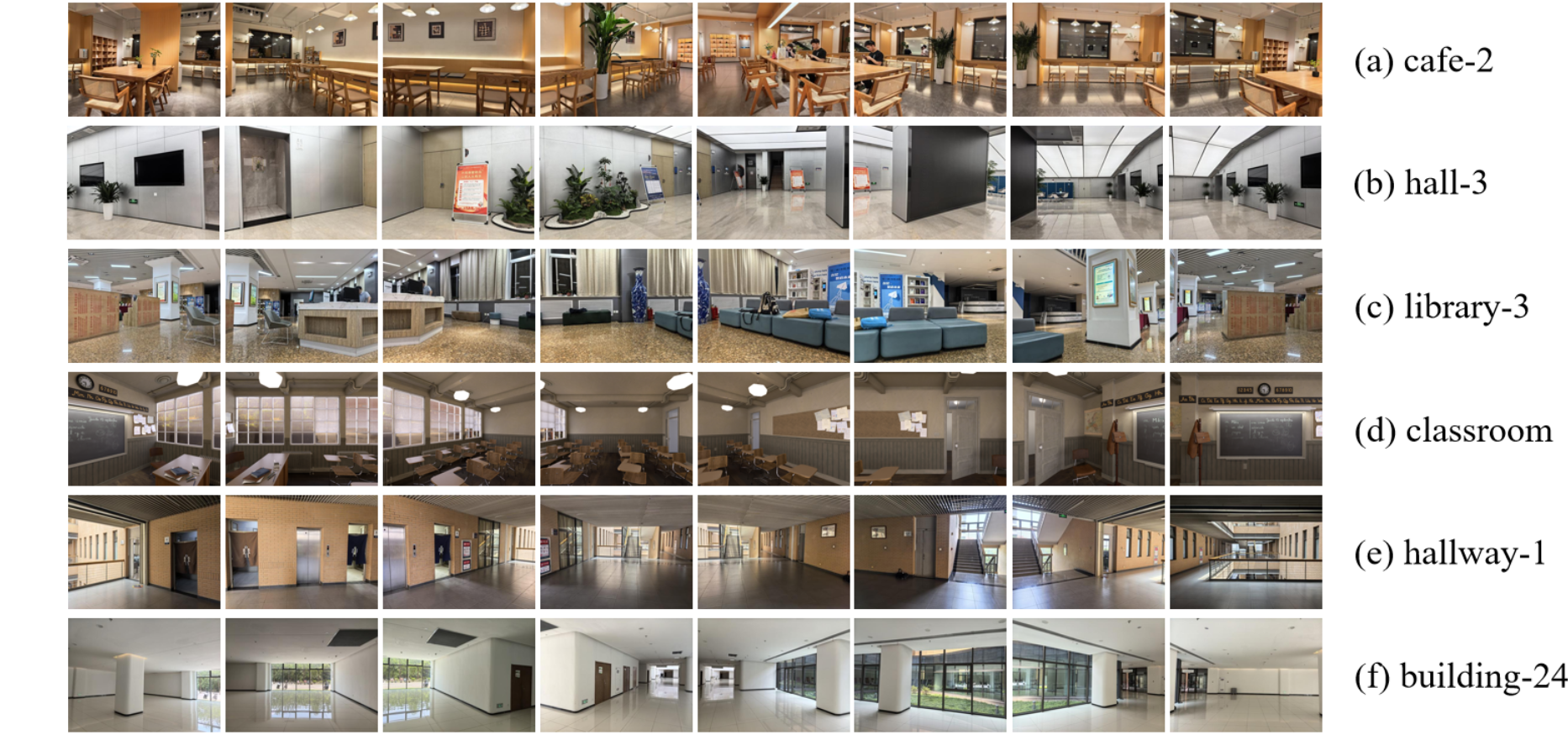}
    \caption{Representative samples from the MVIS dataset. Panels (a)–(d) depict challenging scenes, whereas panels (e)–(f) illustrate simpler scenes.}
    \label{fig:dataset}
\end{figure*}

The MVIS dataset comprises a total of 50 indoor panoramic image groups. By forming pairs of images within each scene, we derived 371 two-image stitching pairs. Detailed information regarding the dataset is presented in Table \ref{tab:dataset}. In instances where multiple panoramic groups exist within a single scene, they are designated using the format “scene name - number” (e.g., “cafe-1”). Scenes are categorized based on the distance between objects and the camera into near-distance challenging scenes and far-distance simple scenes. Near-distance challenging scenes are defined by camera-to-wall distances ranging from 1 to 5 meters, whereas far-distance simple scenes have distances between 5 and 10 meters. Simple scenes generally feature open environments with few walls proximate to the camera and lack distinct nearby objects, as exemplified in Fig. \ref{fig:dataset}(e) and (f). Conversely, challenging scenes typically contain both near and distant objects; for instance, the cafe-2 scene shown in Fig. \ref{fig:dataset}(a) includes chairs close to the camera that introduce significant parallax challenges for stitching. It is important to note that the terms “near” and “far” are relative, as all data were collected indoors, and thus all scenes exhibit some degree of parallax complexity.

\begin{table}[htbp]
\centering
\caption{Comprehensive details of the MVIS dataset}
\label{tab:dataset}
\begin{adjustbox}{max width=\textwidth}
\begin{tabular}{ccccc}
\toprule
\textbf{Category} & \textbf{Scene Name} & \textbf{\makecell{Number of \\Groups}} & \textbf{Difficulty} & \textbf{Baseline (m)} \\
\midrule
\multirow{3}{*}{\makecell{Virtual \\Scenes}} & classroom & 1 & \multirow{3}{*}{Simple} & \multirow{3}{*}{0.1} \\
& spark & 1 & & \\
& harmony & 1 & & \\
\midrule
\multirow{7}{*}{\makecell{Real \\Scenes}} & building & 26 & \multirow{2}{*}{Challenging} & \multirow{2}{*}{0.2} \\
& hallway & 3 & & \\
\cmidrule{2-5}
& cafe & 6 & \multirow{5}{*}{Simple} & \multirow{5}{*}{0.2} \\
& hall & 5 & & \\
& library & 5 & & \\
& dorm & 1 & & \\
& lobby & 1 & & \\
\bottomrule
\end{tabular}
\end{adjustbox}
\end{table}

\begin{table*}[htbp]
\centering
\caption{Performance comparison of various methods on the MVIS dataset. The best results are highlighted in bold red, while the second-best results are underlined in blue.}
\label{tab:performance}
\begin{tabular}{lccccccccc}
\toprule
 & \multicolumn{3}{c}{SSIM $\uparrow$} & \multicolumn{3}{c}{PSNR $\uparrow$} & \multicolumn{3}{c}{LPIPS $\downarrow$} \\
\cmidrule(lr){2-4} \cmidrule(lr){5-7} \cmidrule(lr){8-10}
Method & Simple & Challenging & Average & Simple & Challenging & Average & Simple & Challenging & Average \\
\midrule
Autostitch\cite{2007Automatic} & 0.722 & 0.608 & 0.665 & 20.403 & 16.598 & 18.500 & \textcolor{blue}{\uline{0.212}} & 0.259 & \textcolor{blue}{\uline{0.235}} \\
AANAP\cite{2015Aanap} & 0.642 & 0.552 & 0.597 & 17.194 & 15.598 & 16.396 & 0.289 & 0.309 & 0.299 \\
GSP\cite{2016gsp} & \textcolor{blue}{\uline{0.750}} & 0.620 & 0.685 & \textcolor{blue}{\uline{21.751}} & 17.266 & \textcolor{blue}{\uline{19.509}} & 0.243 & 0.266 & 0.255 \\
UDIS++\cite{2023udis++} & 0.749 & \textcolor{blue}{\uline{0.666}} & \textcolor{blue}{\uline{0.708}} & 20.215 & \textcolor{blue}{\uline{18.108}} & 19.162 & 0.240 & \textcolor{blue}{\uline{0.239}} & 0.240 \\
MHW\cite{2024Parallax} & 0.684 & 0.594 & 0.639 & 18.017 & 16.480 & 17.249 & 0.263 & 0.298 & 0.280 \\
Proposed Method & \textbf{\textcolor{red}{0.770}} & \textbf{\textcolor{red}{0.695}} & \textbf{\textcolor{red}{0.732}} & \textbf{\textcolor{red}{22.657}} & \textbf{\textcolor{red}{18.948}} & \textbf{\textcolor{red}{20.802}} & \textbf{\textcolor{red}{0.182}} & \textbf{\textcolor{red}{0.213}} & \textbf{\textcolor{red}{0.197}} \\
\bottomrule
\end{tabular}
\end{table*}

\subsection{Evaluation of Algorithmic Performance}
\subsubsection{Comparison of Two-Image Stitching Outcomes}

\textbf{Quantitative Metric Analysis.} We conducted a comparative evaluation of our proposed method against representative traditional algorithms—namely Autostitch \cite{2007Automatic}, AANAP \cite{2015Aanap}, and GSP \cite{2016gsp}—as well as recent deep learning-based approaches including UDIS++ \cite{2023udis++} and MHW \cite{2024Parallax}. These methods commonly utilize pairs of images as input for stitching tasks. To assess performance on two-image stitching, we employed these algorithms as baselines and tested them on our bespoke dataset using three established metrics: Peak Signal-to-Noise Ratio (PSNR), Structural Similarity Index Measure (SSIM), and Learned Perceptual Image Patch Similarity (LPIPS) \cite{2018lpips}. For metric computation, we extracted the overlapping regions from the two warped images to evaluate similarity within these shared areas. To ensure fairness, all baseline methods were executed using the original authors’ publicly available code with default parameters, and a uniform mean fusion strategy was applied to mitigate the influence of fusion techniques on the results. The quantitative outcomes are summarized in Table \ref{tab:performance}. Our method demonstrates an average improvement of 1.293 in PSNR, 0.024 in SSIM, and a decrease of 0.038 in LPIPS relative to the best-performing baseline, indicating substantial advantages in scenarios characterized by pronounced depth variation.

\begin{figure*}[htbp]
\centering
\subfloat[Original Images]{
		\includegraphics[scale=0.32]{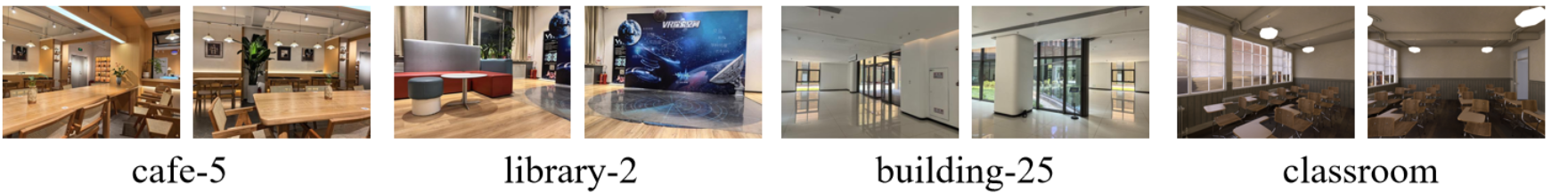}}\\
\subfloat[Autostitch \cite{2007Automatic}]{
		\includegraphics[scale=0.32]{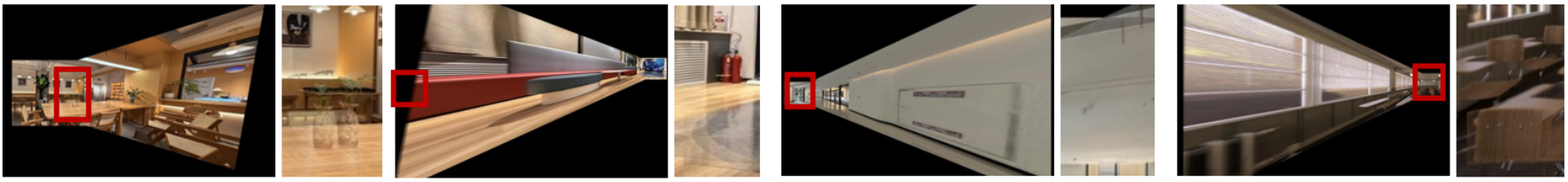}}\\
\subfloat[AANAP \cite{2015Aanap}]{
		\includegraphics[scale=0.32]{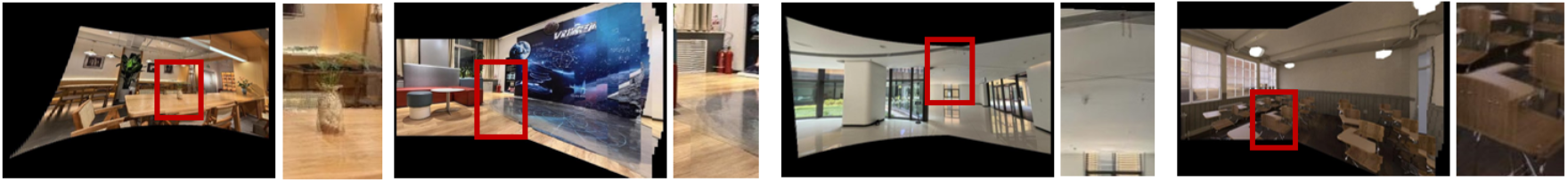}}\\
\subfloat[GSP \cite{2016gsp}]{
		\includegraphics[scale=0.32]{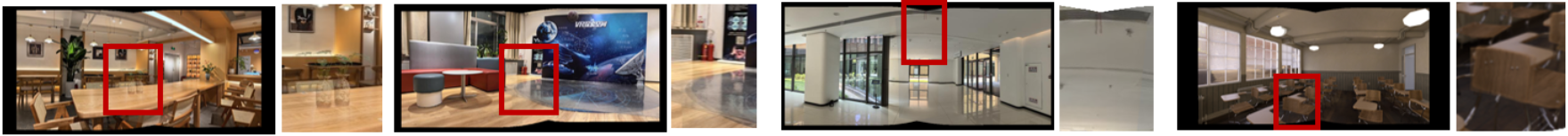}}\\
\subfloat[UDIS++ \cite{2023udis++}]{
		\includegraphics[scale=0.32]{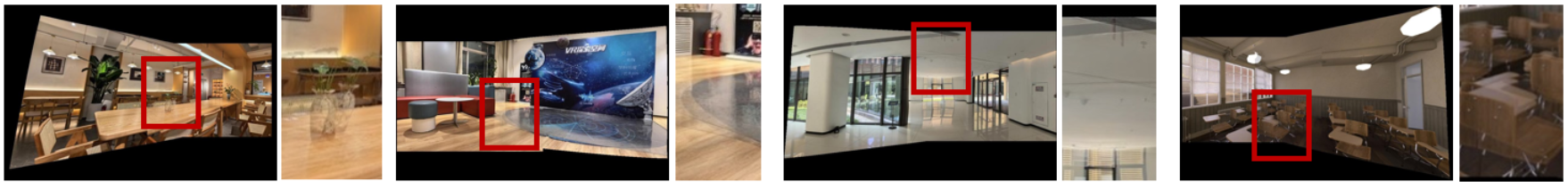}}\\
\subfloat[MHW \cite{2024Parallax}]{
		\includegraphics[scale=0.32]{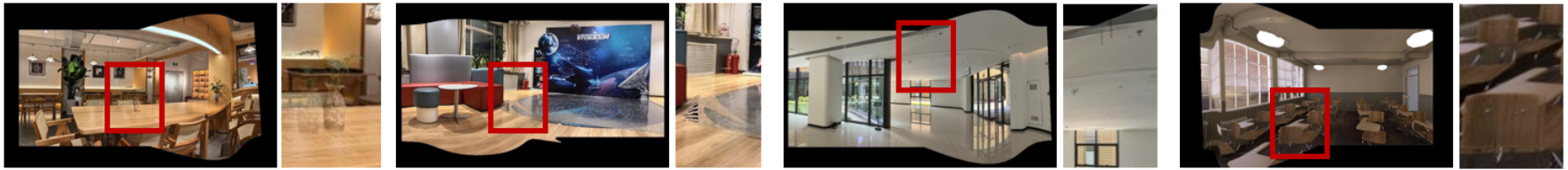}}\\
\subfloat[Proposed Method]{
		\includegraphics[scale=0.32]{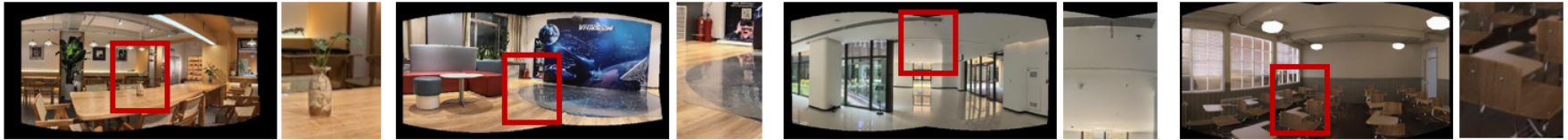}}\\
\caption{Comparison of stitching results in two-image scenarios on the MVIS dataset. Representative overlapping regions were cropped and magnified to illustrate alignment quality. Achieving simultaneous alignment of both near and distant objects in scenes with rich foreground and background content presents significant challenges. Autostitch \cite{2007Automatic} relies exclusively on a single homography matrix for global alignment, lacking local alignment capabilities. AANAP \cite{2015Aanap} employs local grid alignment, which improves local alignment relative to Autostitch but remains inadequate for aligning near and far objects in high-parallax scenes. GSP \cite{2016gsp} further refines local alignment but exhibits instability under large parallax conditions. Deep learning-based methods UDIS++ \cite{2023udis++} and MHW \cite{2024Parallax} inherit the traditional paradigm of combining global and local alignment, yet still struggle to simultaneously align objects at varying depths in scenes with substantial parallax. Our approach reconstructs scene objects via dimensionality lifting and performs separate alignment of objects at different depths within 3D space, thereby effectively addressing the alignment of near and far objects in large parallax scenarios.}
	\label{fig:compare}
\end{figure*}

\textbf{Qualitative Assessment of Stitching Quality.} Figure \ref{fig:compare} illustrates the performance of the aforementioned algorithms in representative two-image stitching scenarios. Given that our custom dataset predominantly comprises indoor close-range scenes, the input images exhibit significant parallax, posing considerable alignment challenges. Evidently, the compared algorithms manifest noticeable artifacts when aligning scene objects. Our method substantially mitigates such artifacts in large parallax contexts by projecting the two-dimensional pixels of each image back into three-dimensional space, thereby enabling separate alignment of objects at distinct depths. For instance, in the final example, the air conditioning vent on the ceiling exhibits pronounced parallax between the two images due to differing viewpoints. While Autostitch \cite{2007Automatic}, AANAP \cite{2015Aanap}, UDIS++ \cite{2023udis++}, and MHW \cite{2024Parallax} fail to align this vent effectively, our method successfully aligns the distant window concurrently with the nearby vent, resulting in superior visual coherence.

\begin{table*}[htbp]
\centering
\caption{Robustness Evaluation of Various Algorithms Across Different Scene Types}
\label{tab:experiment_results}
\begin{adjustbox}{max width=\textwidth}
\begin{tabular}{llcccccc}
\toprule
\textbf{Scene Type} & \textbf{Metric} & \textbf{Autostitch} & \textbf{AANAP} & \textbf{GSP} & \textbf{UDIS++} & \textbf{MHW} & \textbf{Proposed Method} \\
\midrule
\multirow{3}{*}{\makecell{\textbf{Close Range} \\(155 sets)}} & Error Count & 37 & 40 & 6 & 2 & 59 & 0 \\
& Failure Count & 36 & 39 & 12 & 48 & 4 & 7 \\
& Success Rate & 51.61\% & 49.03\% & 88.39\% & 67.74\% & 59.35\% & 95.48\% \\
\midrule
\multirow{3}{*}{\makecell{\textbf{Long Range} \\(216 sets)}} & Error Count & 56 & 71 & 12 & 0 & 80 & 0 \\
& Failure Count & 46 & 16 & 3 & 68 & 3 & 10 \\
& Success Rate & 52.78\% & 59.72\% & 93.06\% & 68.52\% & 61.57\% & 95.37\% \\
\bottomrule
\end{tabular}
\end{adjustbox}
\end{table*}

\begin{figure*}[htbp]
\centering
\subfloat[Autostitch]{
		\includegraphics[scale=0.34]{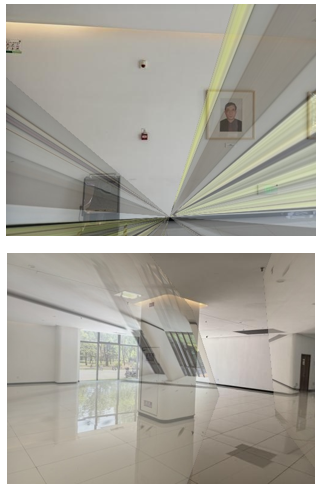}}
\subfloat[UDIS++]{
		\includegraphics[scale=0.34]{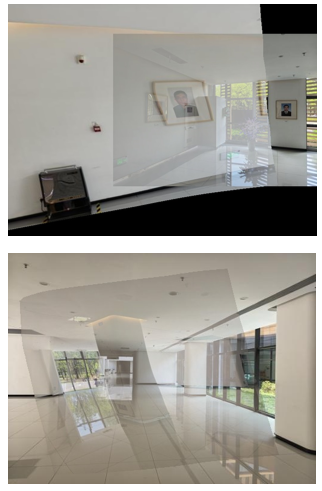}}
\subfloat[GSP]{
		\includegraphics[scale=0.34]{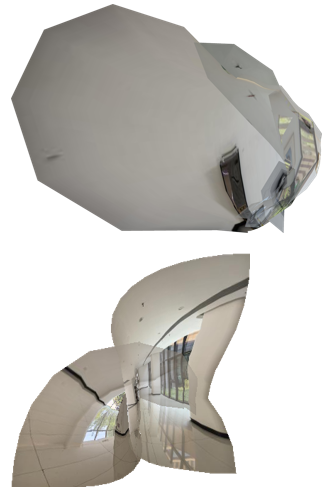}}
\subfloat[Proposed Method]{
		\includegraphics[scale=0.36]{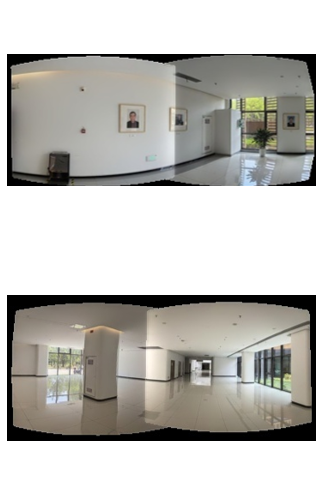}}
\caption{Results on challenging samples. The outcomes of Autostitch \cite{2007Automatic}, UDIS++ \cite{2023udis++}, GSP \cite{2016gsp}, and our method are presented for two distinct scenes. The first three methods failed to produce successful stitching, whereas AANAP \cite{2015Aanap} and MHW \cite{2024Parallax} encountered program crashes during stitching and yielded no output.}
	\label{fig:rubo}
\end{figure*}

\textbf{Robustness Analysis.} The efficacy of image stitching algorithms is closely related to the parallax between input images, which is predominantly influenced by the spatial distance between the camera and scene objects. In close-range scenarios, even minor baselines between shots can induce substantial viewpoint disparities for identical objects, thereby diminishing the robustness of certain stitching methods under large parallax conditions. To investigate algorithmic robustness, we conducted experiments summarized in Table \ref{tab:experiment_results}. The MVIS dataset comprises 371 two-image stitching instances, partitioned into 155 close-range and 216 long-range scenes. Here, “Error Count” denotes the number of program crashes, while “Failure Count” indicates the number of unsuccessful stitching attempts. The data reveal generally lower success rates in close-range scenes due to increased parallax. Notably, GSP \cite{2016gsp} incorporates specific optimizations for 2D and 3D point alignment under large parallax, whereas our method reconstructs the 3D scene through dimensionality lifting, rendering it largely invariant to parallax effects. Consequently, our approach achieves markedly higher success rates compared to the other evaluated methods. Figure \ref{fig:rubo} illustrates stitching results on challenging samples: Autostitch \cite{2007Automatic}, UDIS++ \cite{2023udis++}, and GSP \cite{2016gsp} failed in scenarios characterized by large parallax (top) or limited overlap (bottom). Meanwhile, AANAP \cite{2015Aanap} and MHW \cite{2024Parallax} experienced program crashes during stitching.

\subsubsection{Comparison of Multi-Image Panorama Stitching Results}

\begin{figure*}[!htb]
	\centering
	\subfloat[Autostitch \cite{2007Automatic}]{
		\begin{minipage}{0.15\linewidth}
			\includegraphics[width=\linewidth]{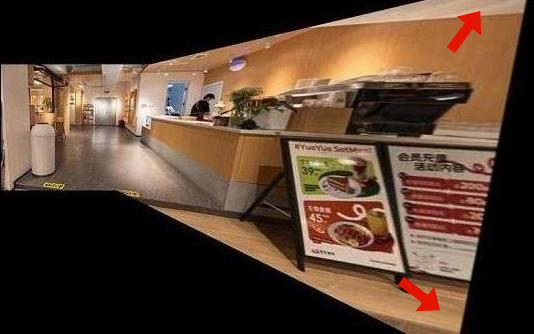}\vspace{3pt}
			\includegraphics[width=\linewidth]{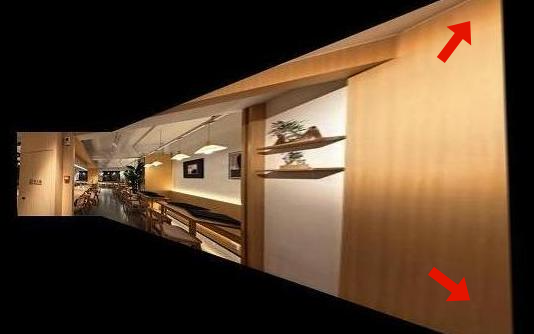}\vspace{3pt}
			\includegraphics[width=\linewidth]{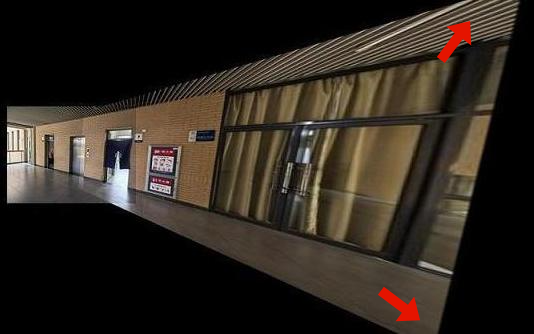}
	\end{minipage}}
	\subfloat[AANAP \cite{2015Aanap}]{
		\begin{minipage}{0.15\linewidth}
			\includegraphics[width=\linewidth]{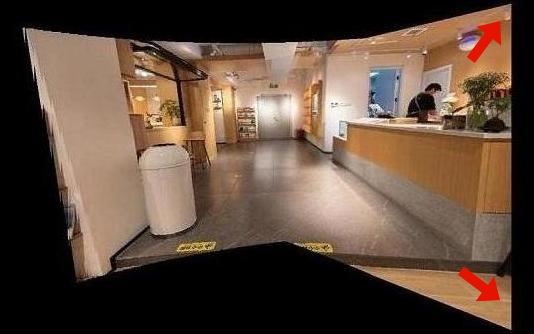}\vspace{3pt}
			\includegraphics[width=\linewidth]{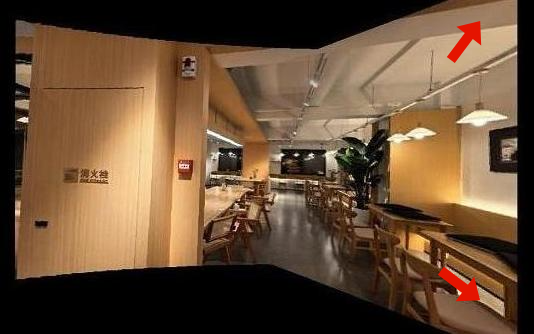}\vspace{3pt}
			\includegraphics[width=\linewidth]{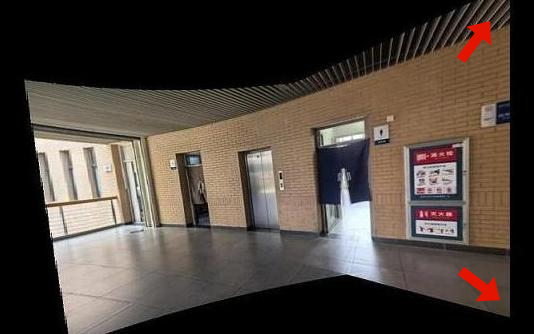}
	\end{minipage}}
	\subfloat[GSP \cite{2016gsp}]{	
		\begin{minipage}{0.15\linewidth}
			\includegraphics[width=\linewidth]{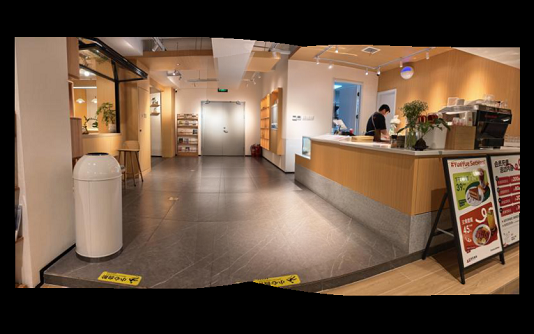}\vspace{3pt}
			\includegraphics[width=\linewidth]{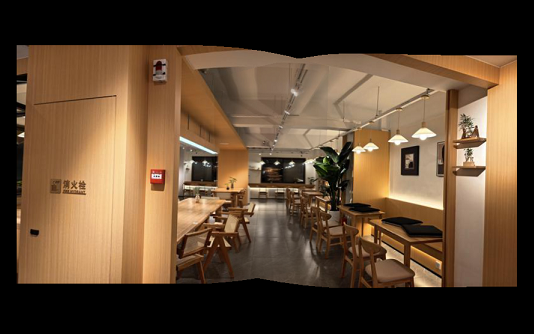}\vspace{3pt}
			\includegraphics[width=\linewidth]{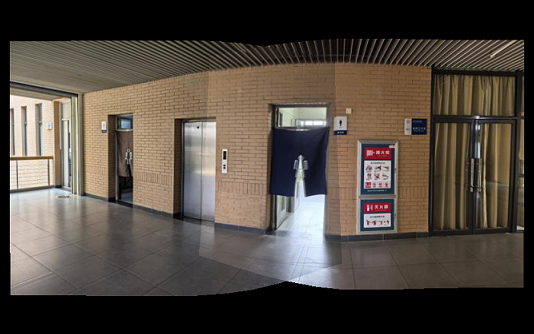}
	\end{minipage}}
 \subfloat[UDIS++ \cite{2023udis++}]{	
		\begin{minipage}{0.15\linewidth}
			\includegraphics[width=\linewidth]{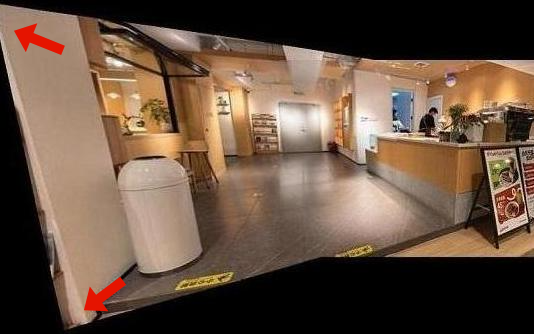}\vspace{3pt}
			\includegraphics[width=\linewidth]{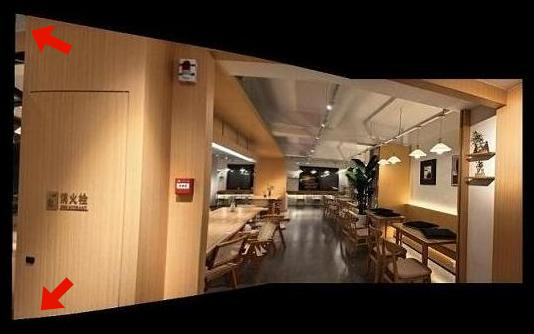}\vspace{3pt}
			\includegraphics[width=\linewidth]{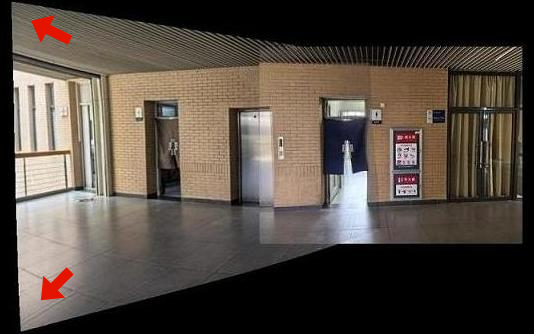}
	\end{minipage}}
	\subfloat[MHW \cite{2024Parallax}]{	
		\begin{minipage}{0.15\linewidth}
			\includegraphics[width=\linewidth]{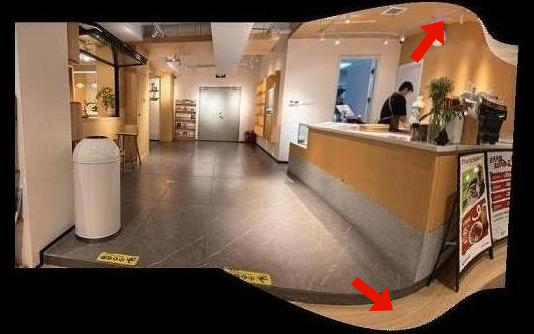}\vspace{3pt}
			\includegraphics[width=\linewidth]{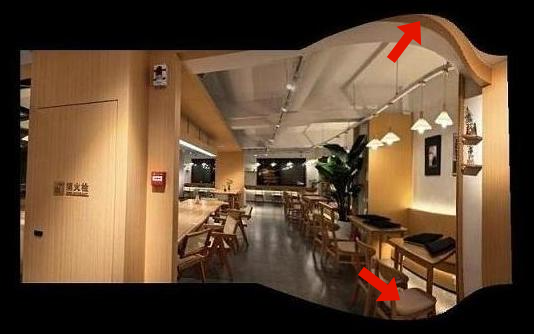}\vspace{3pt}
			\includegraphics[width=\linewidth]{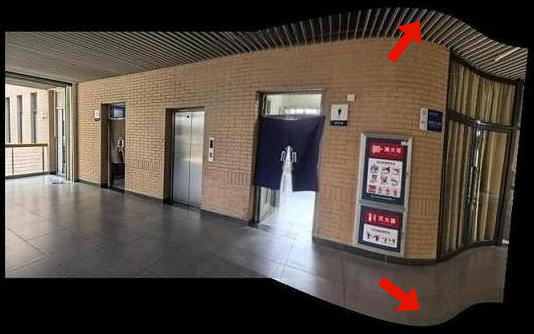}
	\end{minipage}}
    \subfloat[Proposed Method]{	
		\begin{minipage}{0.15\linewidth}
			\includegraphics[width=\linewidth]{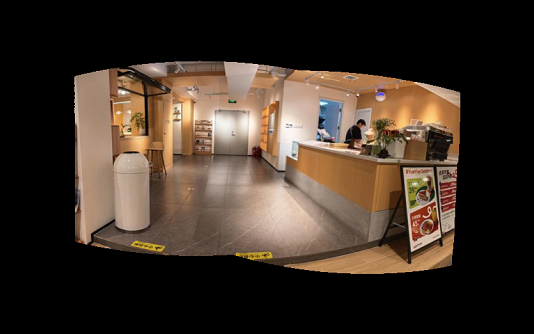}\vspace{3pt}
			\includegraphics[width=\linewidth]{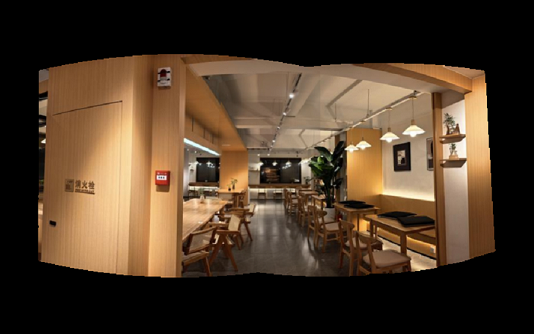}\vspace{3pt}
			\includegraphics[width=\linewidth]{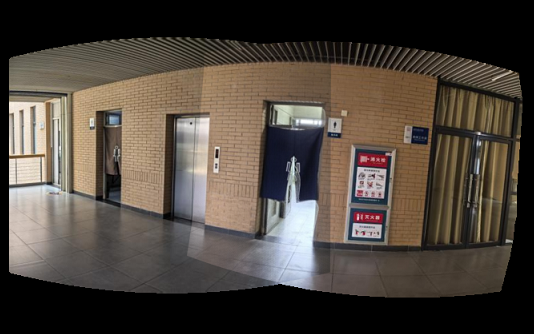}
	\end{minipage}}
	\caption{Comparison of projection distortion effects in two-image stitching scenarios on the MVIS dataset. Complete two-image stitching results are presented. Autostitch \cite{2007Automatic}, AANAP \cite{2015Aanap}, UDIS++ \cite{2023udis++}, and MHW \cite{2024Parallax} exhibit evident distortion along the indicated directions.} 
    \label{fig:project-multi}
\end{figure*}

We further evaluated the panorama stitching capabilities of all compared algorithms in multi-image scenarios. Given that the MVIS dataset predominantly features scenes with substantial parallax, effective stitching methods must exhibit strong resilience to projection distortion. As depicted in Figure \ref{fig:project-multi}, traditional homography-based approaches produce pronounced stretching distortions in non-overlapping regions. Specifically, Autostitch \cite{2007Automatic}, AANAP \cite{2015Aanap}, UDIS++ \cite{2023udis++}, and MHW \cite{2024Parallax} all demonstrate significant distortion along the directions indicated by arrows. When incrementally adding images, these distorted regions complicate feature matching and projection matrix estimation for subsequent images, often culminating in failure due to accumulated distortion. In contrast, our method and GSP \cite{2016gsp} employ cylindrical projection to mitigate projection distortion, thereby preserving geometric consistency even when stitching numerous images.

\begin{figure*}[htbp]
\centering
\subfloat[GSP]{
		\includegraphics[scale=0.29]{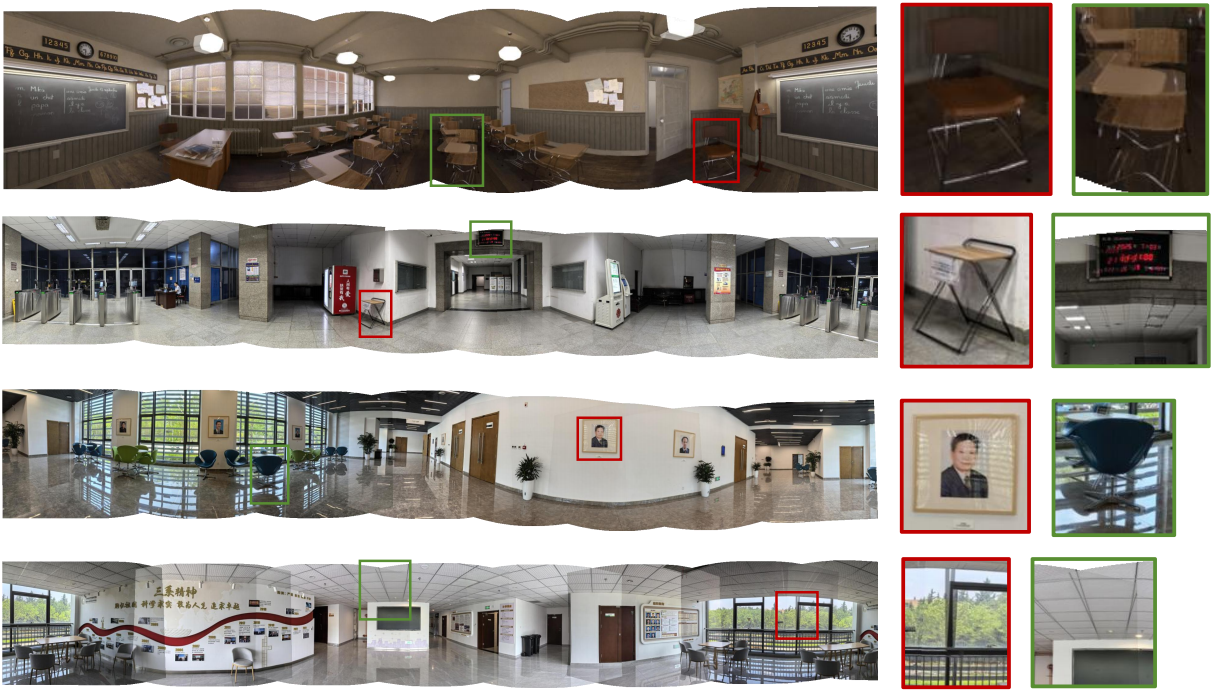}}
\subfloat[Proposed Method]{
		\includegraphics[scale=0.27]{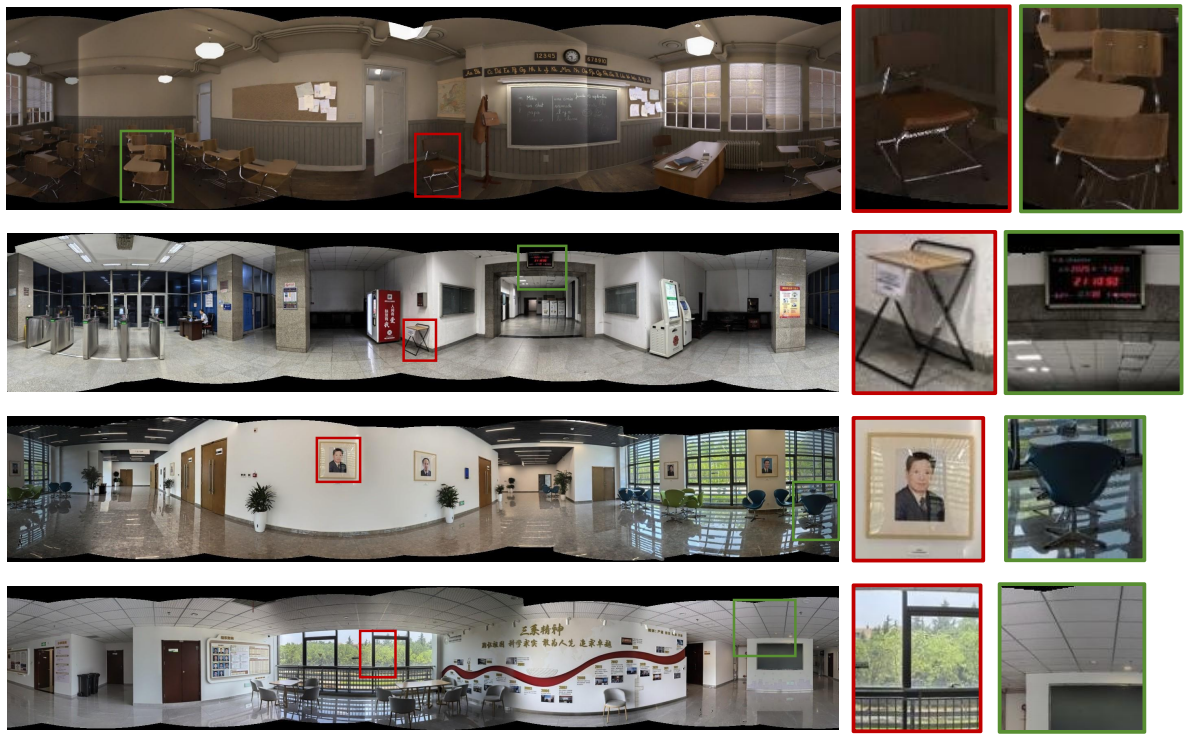}}\\
\caption{Comparison of panorama stitching results. The stitched panorama images are shown alongside magnified views of selected regions. Panel (a) presents results from GSP \cite{2016gsp}, while panel (b) displays outcomes from the proposed method.}
	\label{fig:pano-cmp}
\end{figure*}

Among the evaluated algorithms, only GSP \cite{2016gsp} successfully completes multi-image panorama stitching. We conducted a comparative analysis of multi-image stitching results between GSP and our method on the dataset. Subjective comparisons are depicted in Figure \ref{fig:pano-cmp}. In scenes characterized by substantial depth variation, our approach achieves superior alignment and artifact reduction. For example, in panel (a), the classroom scene contains densely arranged desks proximal to the camera, resulting in significant parallax across different viewpoints. GSP \cite{2016gsp} is unable to simultaneously preserve alignment and shape integrity of objects at varying depths. Conversely, our method obviates concerns regarding distortion by independently segmenting and aligning objects at distinct depths, thereby delivering enhanced visual quality in large parallax panorama stitching tasks. Furthermore, GSP \cite{2016gsp} lacks a loop closure constraint for panoramic stitching, precluding seamless closure between the first and last images.

\subsection{Comparative Experiments}

Within the image stitching framework proposed in this study, both the projection technique and the hole-filling strategy exert notable influences on the quality of the final stitched images. To systematically assess the impact of these two components, we conducted a series of comparative experiments examining different methodological variants and analyzed how these approaches affect the resultant image quality.

\subsubsection{Comparison between Cylindrical Projection and Equidistant Cylindrical Projection}

\begin{figure}[htbp]
\centering
\subfloat[classroom]{
		\includegraphics[scale=0.22]{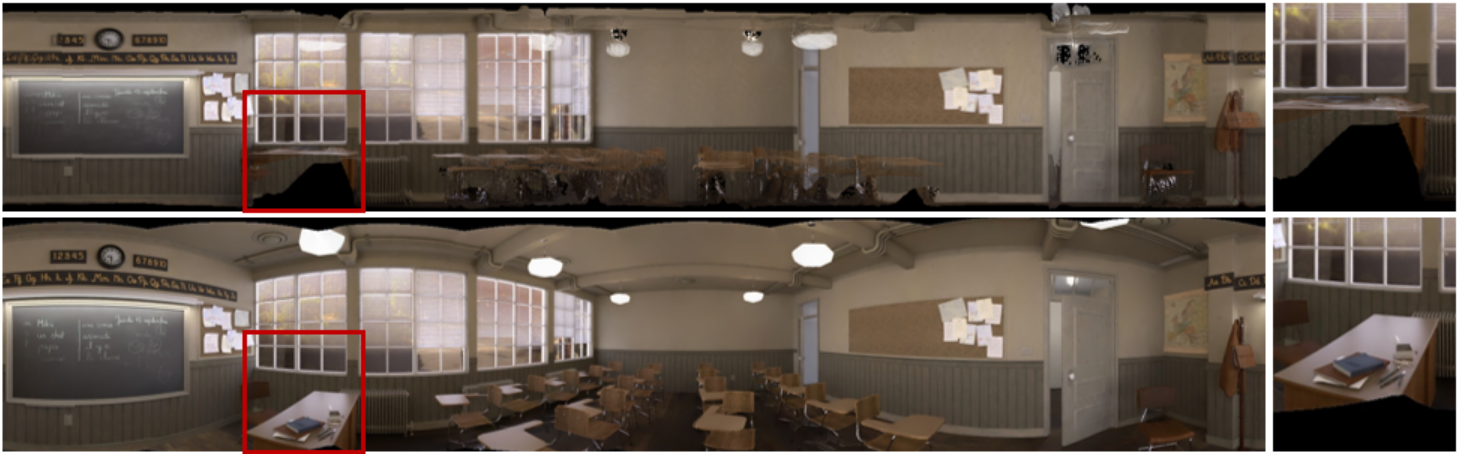}}\vspace{1pt}
\subfloat[hall-1]{
		\includegraphics[scale=0.22]{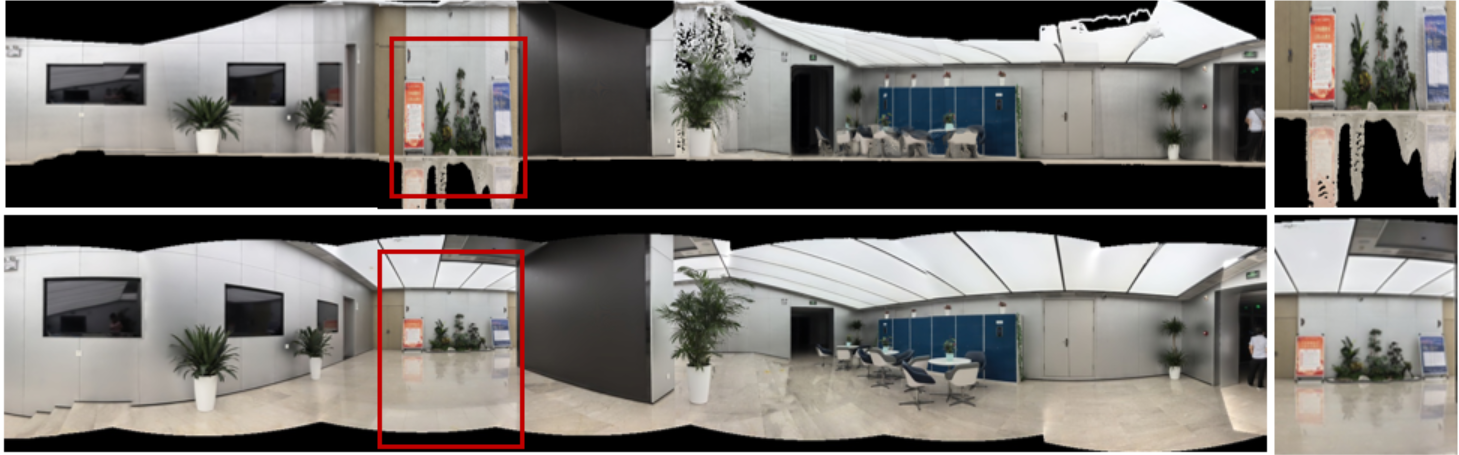}}\\
\subfloat[building-7]{
		\includegraphics[scale=0.22]{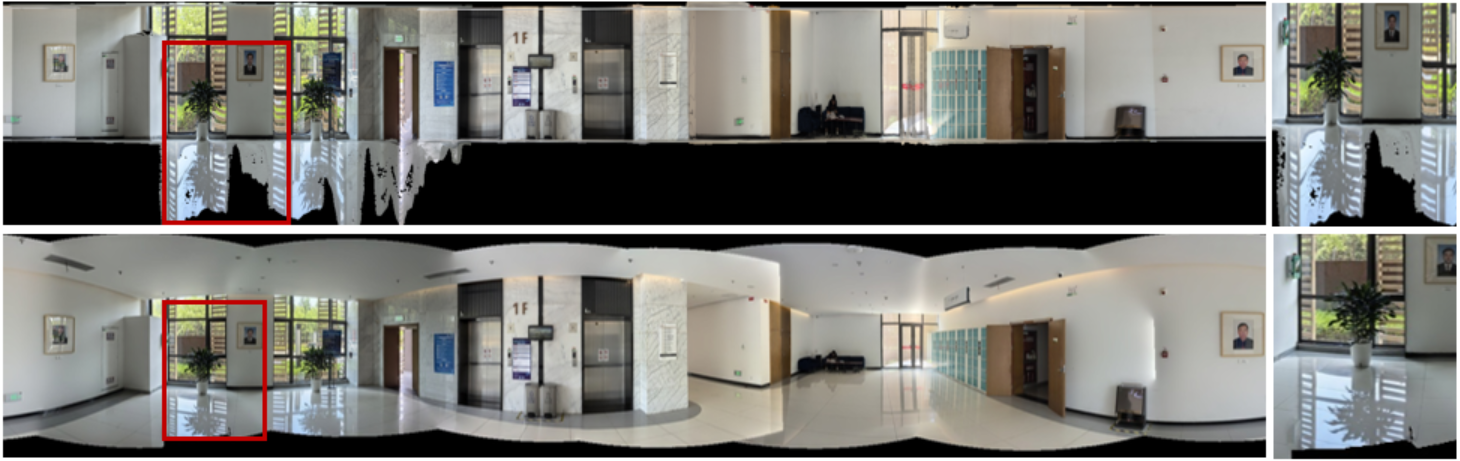}}\vspace{1pt}
\subfloat[building-22]{
		\includegraphics[scale=0.22]{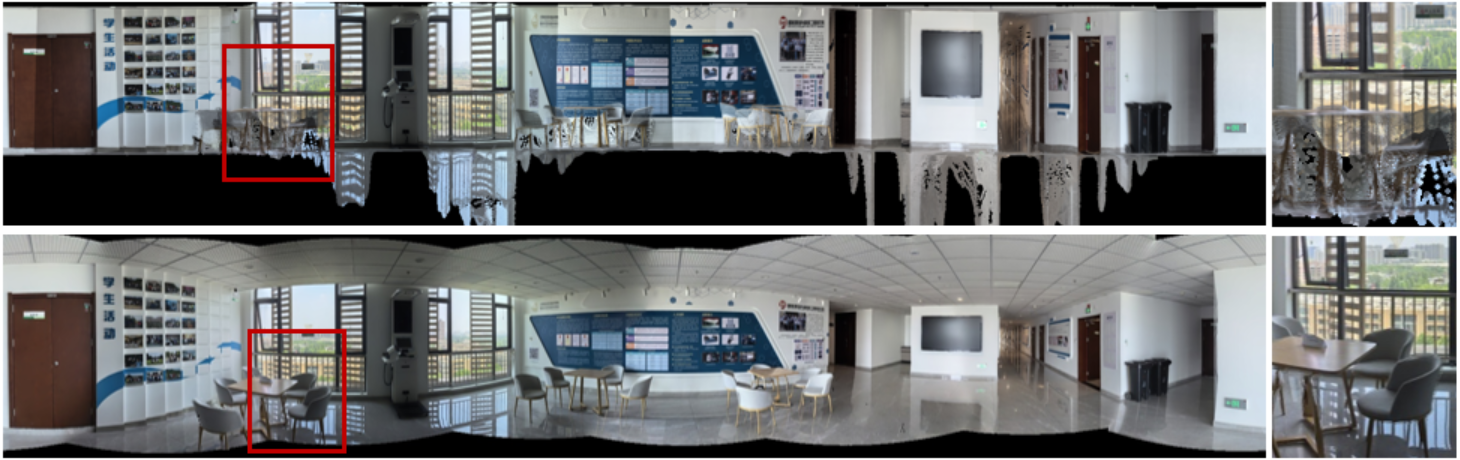}}\\
\caption{Comparison of projection methods across various scenes. In each pair, the lower image depicts the equidistant cylindrical projection result, while the upper image shows the cylindrical projection result. The equidistant cylindrical projection, being more consistent with the original image capture geometry, produces more natural projections with fewer large holes. In contrast, the cylindrical projection projects points parallelly onto the cylinder surface at corresponding heights and views the point cloud from an altered angle, leading to prominent holes around certain objects.}
\label{fig:project-cmp}
\end{figure}

As illustrated in the processing pipeline depicted in Figure \ref{fig:flow}, we projected the dimensionally enhanced point clouds onto the canvas employing two distinct projection methods: cylindrical projection and equidistant cylindrical projection. All other stages of the processing pipeline were held constant. Panoramic stitching was performed on the entire MVIS dataset, and representative results from the classroom, hall1, building7, and building22 scenes were analyzed. The overlapping regions of the stitching outcomes are presented in Figure \ref{fig:project-cmp}, where, in each subfigure, the upper image corresponds to the cylindrical projection result, and the lower image corresponds to the equidistant cylindrical projection utilized in this work.

The comparative analysis reveals that the equidistant cylindrical projection markedly mitigates projection-induced holes caused by occlusions. This advantage arises from its alignment with the original image acquisition geometry. Specifically, under the equidistant cylindrical projection, the source images are composed of rays emanating from a single optical center, closely approximating the original camera viewpoint. Consequently, the occlusion relationships among objects within the point cloud—generated through dimensional elevation from the original viewpoint—are consistent with the viewing ray vectors. Conversely, the cylindrical projection method observes the point cloud from viewpoints orthogonal to the cylinder surface at varying heights along the cylinder axis. This shift in the projection center alters the original viewing angle to an indeterminate one, thereby exposing regions that should be occluded and resulting in the formation of holes. For instance, as demonstrated in Figure \ref{fig:project-cmp}(a) for the classroom scene, the cylindrical projection reveals viewpoints absent in the original images, such as the lateral side of a desk. Since the original point cloud lacks data for these regions, the projection yields substantial holes or missing objects.

Furthermore, during the three-dimensional reconstruction process, the relative heights of objects are preserved with high fidelity. This preservation causes the ground and object reflections to be projected at their authentic heights in the cylindrical projection. For example, in the scene depicted in Figure \ref{fig:project-cmp}(b), the ground—which is originally level—is projected onto a single line, resulting in a loss of ground detail. Simultaneously, the reflection of the exhibition rack on the ground, which is situated at a lower height, is separated from the ground in the cylindrical projection, contradicting typical visual perception.

\subsubsection{Comparison between Interpolation and MAE-Based Hole-Filling Methods}

\begin{figure*}[htbp]
\centering
\subfloat[library-4]{
		\includegraphics[scale=0.3]{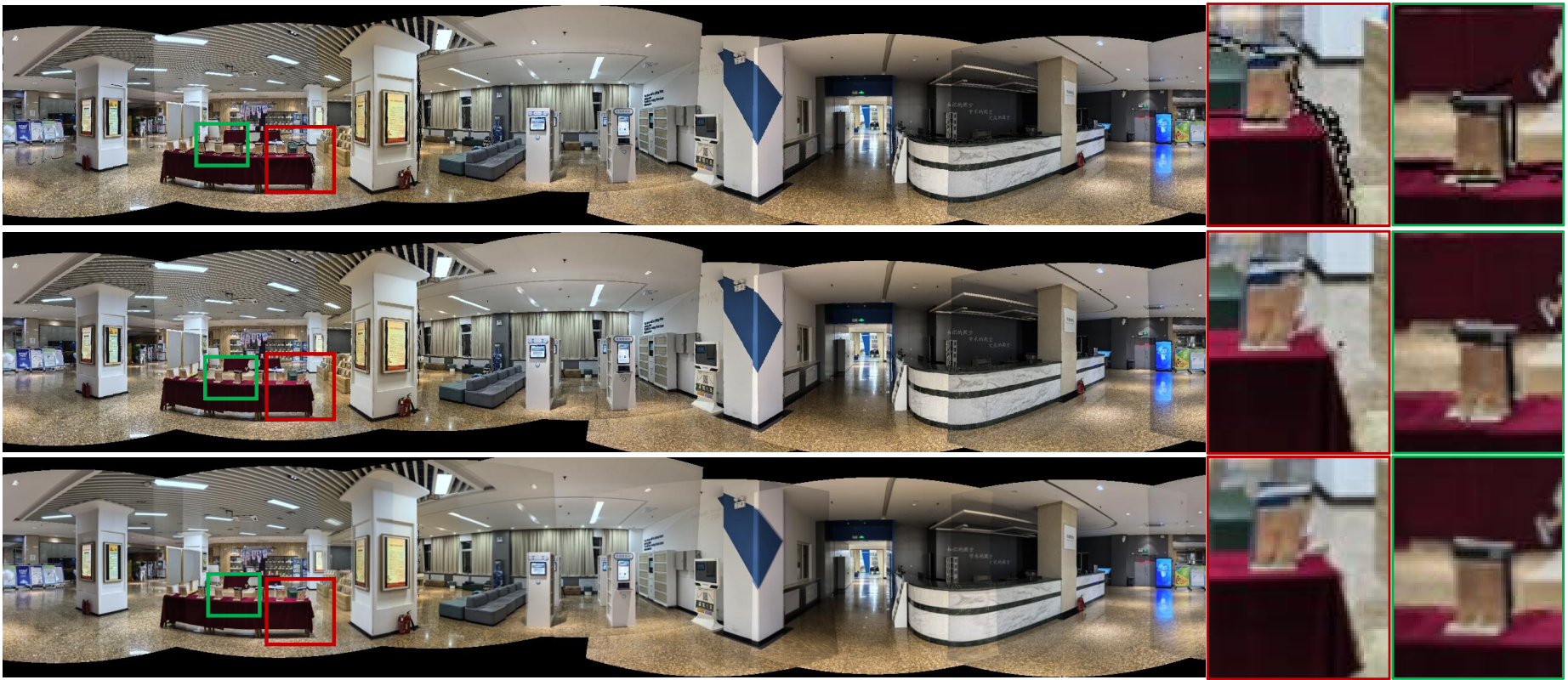}}\\
\subfloat[classroom]{
		\includegraphics[scale=0.3]{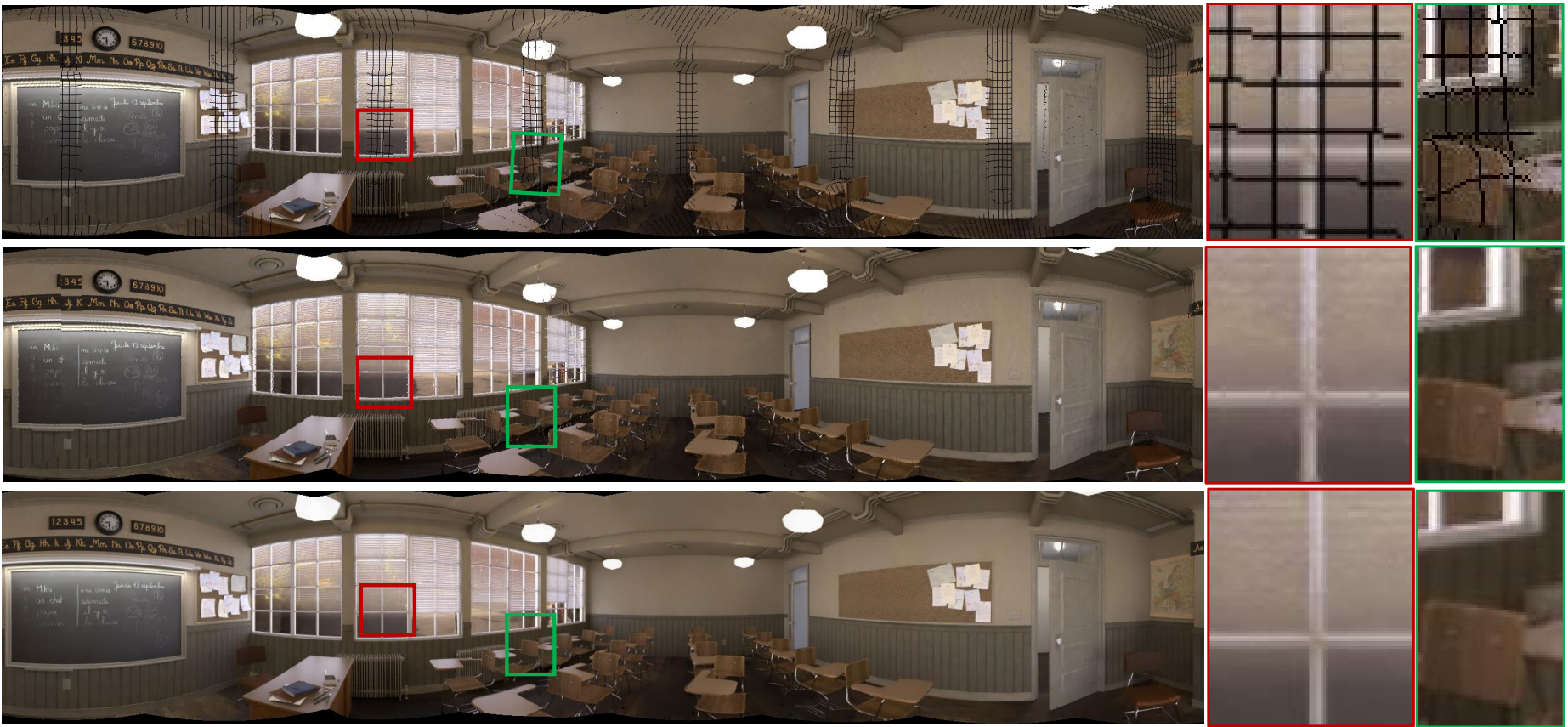}}
\caption{Comparison of hole-filling techniques: interpolation versus MAE-FAR. Within each group, the first image displays the original with black regions indicating holes; the second image shows the result of interpolation-based hole filling; the third image presents the outcome of MAE-FAR hole filling. Insets provide zoomed views of details for enhanced comparison.}
	\label{fig:mae-cmp}
\end{figure*}

Addressing holes in stitched images constitutes a fundamental challenge in stitching computations. Conventional approaches typically perform hole filling in two-dimensional space via backward mapping, effectively employing interpolation algorithms. However, interpolation methods generate only blurred transitional regions based on limited contextual information and are incapable of accurately restoring the geometric boundaries of objects. Although the MAE-FAR method incurs higher computational costs, it leverages a hybrid masking strategy alongside a prior feature fusion mechanism to reconstruct semantic content within missing regions of varying sizes with greater precision. To evaluate the differences in hole-filling quality, we compared interpolation-based results with those obtained through the generative completion approach. The outcomes are illustrated in Figure \ref{fig:mae-cmp}. The first row depicts images following dimensional elevation and projection; the second row shows results after bilinear interpolation hole filling; the third row presents results after deep completion. Selected local areas are magnified in the fourth column to facilitate detailed comparison.

The visual evidence indicates that interpolation-based hole filling computes inserted pixel values based on all surrounding pixels, resulting in blurred boundaries at object edges, such as the edges of desks in (a) and window frames in (b). In contrast, MAE-FAR, which relies on MAE’s prior inference of the overall geometric structure of objects, better preserves object edges and yields superior visual quality.

The disparity in performance between these two hole-filling methods primarily stems from their differing mechanisms for addressing holes. Interpolation assumes local geometric continuity, an assumption that fails at object boundaries where holes commonly occur. Conversely, MAE, through extensive pretraining, acquires global semantic priors and comprehends larger-scale structural relationships within scenes, enabling it to generate plausible completions even in the presence of substantial missing regions. In stitching scenarios, holes typically arise from discontinuities at object boundaries in three-dimensional space, violating the continuity assumption underpinning interpolation. Consequently, interpolation-based methods frequently fail to achieve satisfactory hole-filling results.

\section{Conclusion}
This study presents a comprehensive examination of the principal challenges in image stitching technology, specifically addressing projection distortion and alignment artifacts encountered in complex scenes characterized by significant parallax. Recognizing the intrinsic geometric constraints of conventional two-dimensional approaches reliant on homography matrices, we introduce an innovative panorama generation framework that synergistically combines dimensionality elevation with dimensionality reduction projection. This framework fundamentally redefines the problem-solving paradigm by mapping the input image sequence into three-dimensional space for subsequent registration and fusion.

The principal contribution of this research is the development of a novel panoramic stitching framework, comprising three critical stages that have been rigorously designed and empirically validated. Initially, the DUSt3R network is utilized to perform unconstrained dense 3D reconstruction, thereby transforming the original 2D images into scene point clouds enriched with accurate depth information. This transformation establishes a geometric basis for the independent and fine-grained alignment of objects situated at varying depths. Subsequently, a tailored equidistant cylindrical projection method, anchored on a specified optical center, is devised to project the fused 3D point cloud back onto a two-dimensional manifold and subsequently unfold this manifold onto a planar canvas. This approach effectively mitigates the non-uniform stretching distortions commonly induced by traditional cylindrical image projections, thereby preserving geometric consistency within the resulting panorama. Finally, to address the issue of holes generated during projection, an enhanced masked autoencoder model is introduced. Employing a hybrid masking strategy alongside a prior feature fusion mechanism, this model facilitates semantically accurate inpainting of extensive missing regions, thereby ensuring the visual coherence of the final output.

When compared to existing methodologies, the proposed approach demonstrates marked advantages in multi-image panoramic stitching scenarios. Given the scarcity of datasets containing multiple images per scene, we constructed the MVIS dataset, which encompasses multiple images providing comprehensive 360-degree coverage for each scene. Both qualitative and quantitative evaluations conducted on the MVIS dataset reveal that our method significantly surpasses mainstream algorithms in terms of registration precision and visual naturalness, successfully achieving distortion-free, high-quality panoramic stitching. These findings offer novel insights and effective solutions for advancing image stitching research in complex environments.

This work primarily focuses on the proposition and validation of a new framework and associated innovative concepts. Future research will aim to enhance the generalizability and computational efficiency of the dimensionality elevation process. Nonetheless, given the rapid progress in three-dimensional spatial computing technologies in recent years, it is anticipated that the quality and computational performance of panoramic stitching based on this framework will experience substantial improvements.

\section*{Acknowledgments}
This should be a simple paragraph before the References to thank those individuals and institutions who have supported your work on this article.

{\small
\bibliographystyle{IEEEtran}
\bibliography{liftProj_bib}

@inproceedings{Kang_1998,   
title={Survey of image-based rendering techniques},  
url={https://doi.org/10.1117/12.333774},  
DOI={10.1117/12.333774},  
booktitle={SPIE Proceedings,Videometrics VI},  
author={Kang, Sing B.},  
year={1998},  month={Dec},  pages={2–16},  language={en-US}  }

@article{2004Distinctive,   
title={Distinctive Image Features from Scale-Invariant Keypoints},  
url={http://dx.doi.org/10.1023/b:visi.0000029664.99615.94},  
DOI={10.1023/b:visi.0000029664.99615.94},  
journal={International Journal of Computer Vision},  
author={Lowe, David G.},  
year={2004},  month={Nov},  pages={91–110},  language={en-US}  }

@article{2007Automatic,   
title={Automatic Panoramic Image Stitching using Invariant Features},  
url={http://dx.doi.org/10.1007/s11263-006-0002-3},  
DOI={10.1007/s11263-006-0002-3},  
journal={International Journal of Computer Vision},  
author={Brown, Matthew and Lowe, David G.},  
year={2007},  month={Aug},  pages={59–73},  language={en-US}  }

@article{2013Apap,   
title={As-Projective-As-Possible Image Stitching with Moving DLT},  
url={http://dx.doi.org/10.1109/tpami.2013.247},  
DOI={10.1109/tpami.2013.247},  
journal={IEEE Transactions on Pattern Analysis and Machine Intelligence},  
author={Zaragoza, JulioH. and Chin, Tat-Jun and Tran, Quoc-Huy and Brown, MichaelS. and Suter, David},  
year={2014},  month={Jul},  pages={1285–1298},  language={en-US}  }

@inproceedings{2014Sphp,   
title={Shape-Preserving Half-Projective Warps for Image Stitching},  
url={http://dx.doi.org/10.1109/cvpr.2014.422},  
DOI={10.1109/cvpr.2014.422},  
booktitle={2014 IEEE Conference on Computer Vision and Pattern Recognition(CVPR)},  
author={Chang, Che-Han and Sato, Yoichi and Chuang, Yung-Yu},  
year={2014},  month={Jun},  language={en-US}  }

@inproceedings{2011DHW,   
title={Constructing image panoramas using dual-homography warping},  
url={http://dx.doi.org/10.1109/cvpr.2011.5995433},  
DOI={10.1109/cvpr.2011.5995433},  
booktitle={2011 IEEE Conference on Computer Vision and Pattern Recognition(CVPR)},  
author={Gao, Junhong and Kim, Seon Joo and Brown, Michael S.},  
year={2011},  month={Jun},  language={en-US}  }

@article{2018ela,   
title={Parallax-Tolerant Image Stitching Based on Robust Elastic Warping},  
url={http://dx.doi.org/10.1109/tmm.2017.2777461},  
DOI={10.1109/tmm.2017.2777461},  
journal={IEEE Transactions on Multimedia},  
author={Li, Jing and Wang, Zhengming and Lai, Shiming and Zhai, Yongping and Zhang, Maojun},  
year={2018},  month={Jul},  pages={1672–1687},  language={en-US}  }

@inproceedings{2015Aanap,   
title={Adaptive as-natural-as-possible image stitching},  
url={http://dx.doi.org/10.1109/cvpr.2015.7298719},  
DOI={10.1109/cvpr.2015.7298719},  
booktitle={2015 IEEE Conference on Computer Vision and Pattern Recognition (CVPR)},  
author={Lin, Chung-Ching and Pankanti, Sharathchandra U. and Ramamurthy, Karthikeyan Natesan and Aravkin, Aleksandr Y.},  
year={2015},  month={Jun},  language={en-US}  }

@inbook{2016gsp,   
title={Natural Image Stitching with the Global Similarity Prior},  
url={http://dx.doi.org/10.1007/978-3-319-46454-1_12},  
DOI={10.1007/978-3-319-46454-1_12},  
booktitle={European Conference on Computer Vision (ECCV)},  
author={Chen, Yu-Sheng and Chuang, Yung-Yu},  
year={2016},  month={Jan},  pages={186–201},  language={en-US}  }

@inproceedings{2022ges,   
title={Geometric Structure Preserving Warp for Natural Image Stitching},
DOI={10.1109/CVPR52688.2022.00367},  
booktitle={2022 IEEE/CVF Conference on Computer Vision and Pattern Recognition (CVPR)},  
author={Du, Peng and Ning, Jifeng and Cui, Jiguang and Huang, Shaoli and Wang, Xinchao and Wang, Jiaxin},  
year={2022},  month={Jun},  language={en-US}  }

@article{2017Image,   
title={Image stitching by line-guided local warping with global similarity constraint},  
url={http://dx.doi.org/10.1016/j.patcog.2018.06.013},  
DOI={10.1016/j.patcog.2018.06.013},  
journal={Pattern Recognition},  
author={Xiang, Tian-Zhu and Xia, Gui-Song and Bai, Xiang and Zhang, Liangpei},  
year={2018},  month={Nov},  pages={481–497},  language={en-US}  }

@inproceedings{2021lpc,   
title={Leveraging Line-point Consistence to Preserve Structures for Wide Parallax Image Stitching},  
url={http://dx.doi.org/10.1109/cvpr46437.2021.01201},  
DOI={10.1109/cvpr46437.2021.01201},  
booktitle={2021 IEEE/CVF Conference on Computer Vision and Pattern Recognition (CVPR)},  
author={Jia, Qi and Li, ZhengJun and Fan, Xin and Zhao, Haotian and Teng, Shiyu and Ye, Xinchen and Latecki, Longin Jan},  
year={2021},  month={Jun},  language={en-US}  }

@article{2018Quasi,   
title={Quasi-Homography Warps in Image Stitching},  
url={http://dx.doi.org/10.1109/tmm.2017.2771566},  
DOI={10.1109/tmm.2017.2771566},  
journal={IEEE Transactions on Multimedia},  
author={Li, Nan and Xu, Yifang and Wang, Chao},  
year={2018},  month={Jun},  pages={1365–1375},  language={en-US}  }

@article{2019spw,   
title={Single-Perspective Warps in Natural Image Stitching},  
url={http://dx.doi.org/10.1109/tip.2019.2934344},  
DOI={10.1109/tip.2019.2934344},  
journal={IEEE Transactions on Image Processing},  
author={Liao, Tianli and Li, Nan},  
year={2019},  month={Jan},  pages={724–735},  language={en-US}  }

@article{2022depmap,
  title={Natural image stitching using depth maps},
  url={https://doi.org/10.1016/j.image.2025.117438},  
  DOI={10.48550/arXiv.2202.06276},  
  author={Liao, Tianli and Li, Nan},
  journal={Signal Processing: Image Communication},
  pages={117438},
  year={2025},
  publisher={Elsevier},  language={en-US}
}

@article{2024Objgsp, 
title={Object-level Geometric Structure Preserving for Natural Image Stitching}, 
volume={39}, 
url={https://ojs.aaai.org/index.php/AAAI/article/view/32188}, 
DOI={10.1609/aaai.v39i2.32188}, 
number={2}, 
journal={Proceedings of the AAAI Conference on Artificial Intelligence}, 
author={Cai, Wenxiao and Yang, Wankou}, 
year={2025}, month={Apr.}, pages={1926-1934} }

@article{2021semplaner,   
title={Image Stitching Based on Semantic Planar Region Consensus},  
url={http://dx.doi.org/10.1109/tip.2021.3086079},  
DOI={10.1109/tip.2021.3086079},  
journal={IEEE Transactions on Image Processing},  
author={Li, Aocheng and Guo, Jie and Guo, Yanwen},  
year={2021},  month={Jan},  pages={5545–5558},  language={en-US}  }

@article{2018deephomo,   
title={Unsupervised Deep Homography: A Fast and Robust Homography Estimation Model},  
url={http://dx.doi.org/10.1109/lra.2018.2809549},  
DOI={10.1109/lra.2018.2809549},  
journal={IEEE Robotics and Automation Letters},  
author={Nguyen, Ty and Chen, Steven W. and Shivakumar, Shreyas S. and Taylor, Camillo Jose and Kumar, Vijay},  
year={2018},  month={Jul},  pages={2346–2353},  language={en-US}  }

@INPROCEEDINGS{2023udis++,
  author={Nie, Lang and Lin, Chunyu and Liao, Kang and Liu, Shuaicheng and Zhao, Yao},
  booktitle={Proceedings of the IEEE/CVF International Conference on Computer Vision (ICCV)}, 
  title={Parallax-Tolerant Unsupervised Deep Image Stitching}, 
  year={2023},
  volume={},
  number={},
  pages={7365-7374},
  doi={10.1109/ICCV51070.2023.00680}}

@inbook{2020Content,   
title={Content-Aware Unsupervised Deep Homography Estimation},  
url={http://dx.doi.org/10.1007/978-3-030-58452-8_38},  
DOI={10.1007/978-3-030-58452-8_38},  
booktitle={European Conference on Computer Vision (ECCV)},  
author={Zhang, Jirong and Wang, Chuan and Liu, Shuaicheng and Jia, Lanpeng and Ye, Nianjin and Wang, Jue and Zhou, Ji and Sun, Jian},  
year={2020},  month={Jan},  pages={653–669},  language={en-US}  }

@article{2021udis,   
title={Unsupervised Deep Image Stitching: Reconstructing Stitched Features to Images},  
url={http://dx.doi.org/10.1109/tip.2021.3092828},  
DOI={10.1109/tip.2021.3092828},  
journal={IEEE Transactions on Image Processing},  
author={Nie, Lang and Lin, Chunyu and Liao, Kang and Liu, Shuaicheng and Zhao, Yao},  
year={2021},  month={Jan},  pages={6184–6197},  language={en-US}  }

@inproceedings{2023pixel,
author = {Jia, Qi and Feng, Xiaomei and Liu, Yu and Fan, Xin and Latecki, Longin Jan},
title = {Learning Pixel-wise Alignment for Unsupervised Image Stitching},
year = {2023},
url = {https://doi.org/10.1145/3581783.3612298},
doi = {10.1145/3581783.3612298},
booktitle = {Proceedings of the 31st ACM International Conference on Multimedia},
pages = {1392–1400}
}

@InProceedings{2025PixelStitch,
    author    = {Jin, Hengzhe and Nie, Lang and Lin, Chunyu and Feng, Xiaomei and Zhao, Yao},
    title     = {PixelStitch: Structure-Preserving Pixel-Wise Bidirectional Warps for Unsupervised Image Stitching},
    booktitle = {Proceedings of the IEEE/CVF International Conference on Computer Vision (ICCV)},
    month     = {October},
    year      = {2025},
    pages     = {28125-28134}
}

@article{2019tfa,   
title={Local-Adaptive Image Alignment Based on Triangular Facet Approximation},  
url={http://dx.doi.org/10.1109/tip.2019.2949424},  
DOI={10.1109/tip.2019.2949424},  
journal={IEEE Transactions on Image Processing},  
author={Li, Jing and Deng, Baosong and Tang, Rongfu and Wang, Zhengming and Yan, Ye},  
year={2020},  month={Jan},  pages={2356–2369},  language={en-US}  }

@inproceedings{2020Wrbis,   
title={Warping Residual Based Image Stitching for Large Parallax},  
url={http://dx.doi.org/10.1109/cvpr42600.2020.00822},  
DOI={10.1109/cvpr42600.2020.00822},  
booktitle={2020 IEEE/CVF Conference on Computer Vision and Pattern Recognition (CVPR)},  
author={Lee, Kyu-Yul and Sim, Jae-Young},  
year={2020},  month={Jun},  language={en-US}  }

@article{2010LSD,   
title={LSD: A Fast Line Segment Detector with a False Detection Control},  
url={http://dx.doi.org/10.1109/tpami.2008.300},  
DOI={10.1109/tpami.2008.300},  
journal={IEEE Transactions on Pattern Analysis and Machine Intelligence},  
author={von Gioi, R.G. and Jakubowicz, J. and Morel, J.-M. and Randall, G.},  
year={2010},  month={Apr},  pages={722–732},  language={en-US}  }

@inproceedings{2015dfw,   
title={Dual-Feature Warping-Based Motion Model Estimation},  
url={http://dx.doi.org/10.1109/iccv.2015.487},  
DOI={10.1109/iccv.2015.487},  
booktitle={Proceedings of the IEEE International Conference on Computer Vision (ICCV)},  
author={Li, Shiwei and Yuan, Lu and Sun, Jian and Quan, Long},  year={2015},  month={Dec},  language={en-US}  }

@article{2020udisd,   
title={A view-free image stitching network based on global homography},  
url={http://dx.doi.org/10.1016/j.jvcir.2020.102950},  
DOI={10.1016/j.jvcir.2020.102950},  
journal={Journal of Visual Communication and Image Representation},  
author={Nie, Lang and Lin, Chunyu and Liao, Kang and Liu, Meiqin and Zhao, Yao},  
year={2020},  month={Nov},  pages={102950},  language={en-US}  }

@article{2021nie,   
title={Learning edge-preserved image stitching from multi-scale deep homography},  
url={http://dx.doi.org/10.1016/j.neucom.2021.12.032},  
DOI={10.1016/j.neucom.2021.12.032},  
journal={Neurocomputing},  
author={Nie, Lang and Lin, Chunyu and Liao, Kang and Zhao, Yao},  
year={2021},  month={Jun},  pages={533–543},  language={en-US}  }

@INPROCEEDINGS{2024bfr,
  author={Kim, Minsu and Lee, Jaewon and Lee, Byeonghun and Im, Sunghoon and Jin, Kyong Hwan},
  booktitle={2024 IEEE/CVF Winter Conference on Applications of Computer Vision (WACV)}, 
  title={Implicit Neural Image Stitching With Enhanced and Blended Feature Reconstruction}, 
  year={2024},
  pages={4075-4084},
  doi={10.1109/WACV57701.2024.00404}}

@INPROCEEDINGS{2024dbc,
  author={Kim, Minsu and Lee, Yongjun and Han, Woo Kyoung and Hwan Jin, Kyong},
  booktitle={2024 IEEE/CVF Winter Conference on Applications of Computer Vision (WACV)}, 
  title={Learning Residual Elastic Warps for Image Stitching under Dirichlet Boundary Condition}, 
  year={2024},
  pages={4004-4012},
  doi={10.1109/WACV57701.2024.00397}}

@article{2024RecStitchNet,
author = {Zhang, Yun and Lai, Yu-Kun and Nie, Lang and Zhang, Fang-Lue and Xu, Lin},
year = {2024},
month = {08},
pages = {687-703},
title = {RecStitchNet: Learning to stitch images with rectangular boundaries},
volume = {10},
journal = {Computational Visual Media},
doi = {10.1007/s41095-024-0420-6}
}

@article{2023pww, 
title={Pixel-Wise Warping for Deep Image Stitching}, 
volume={37}, 
url={https://ojs.aaai.org/index.php/AAAI/article/view/25202}, 
DOI={10.1609/aaai.v37i1.25202}, 
number={1}, 
journal={Proceedings of the AAAI Conference on Artificial Intelligence}, 
author={Kweon, Hyeokjun and Kim, Hyeonseong and Kang, Yoonsu and Yoon, Youngho and Jeong, WooSeong and Yoon, Kuk-Jin}, 
year={2023}, month={Jun.}, pages={1196-1204} }

@misc{2024dust3r,
      title={DUSt3R: Geometric 3D Vision Made Easy}, 
      author={Shuzhe Wang and Vincent Leroy and Yohann Cabon and Boris Chidlovskii and Jerome Revaud},
      year={2024},
      eprint={2312.14132},
      archivePrefix={arXiv},
      primaryClass={cs.CV},
      url={https://arxiv.org/abs/2312.14132}, 
}

@article{2024Parallax,
title = {Parallax-tolerant image stitching via segmentation-guided multi-homography warping},
journal = {Signal Processing},
volume = {230},
pages = {109860},
year = {2025},
issn = {0165-1684},
doi = {https://doi.org/10.1016/j.sigpro.2024.109860},
url = {https://www.sciencedirect.com/science/article/pii/S0165168424004808},
author = {Tianli Liao and Ce Wang and Lei Li and Guangen Liu and Nan Li}
}

@inproceedings{2018lpips,   
title={The Unreasonable Effectiveness of Deep Features as a Perceptual Metric},  
url={http://dx.doi.org/10.1109/cvpr.2018.00068},  
DOI={10.1109/cvpr.2018.00068},  
booktitle={2018 IEEE/CVF Conference on Computer Vision and Pattern Recognition},  
author={Zhang, Richard and Isola, Phillip and Efros, Alexei A. and Shechtman, Eli and Wang, Oliver},  
year={2018},  month={Jun},  language={en-US}  }

@inproceedings{wang2025vggt,
  title={VGGT: Visual Geometry Grounded Transformer},
  author={Wang, Jianyuan and Chen, Minghao and Karaev, Nikita and Vedaldi, Andrea and Rupprecht, Christian and Novotny, David},
  booktitle={Proceedings of the IEEE/CVF Conference on Computer Vision and Pattern Recognition(CVPR)},
  year={2025}
}

@article{2005The,   
title={The geometric error for homographies},  
url={http://dx.doi.org/10.1016/j.cviu.2004.03.004},  
DOI={10.1016/j.cviu.2004.03.004},  
journal={Computer Vision and Image Understanding},  
author={Chum, Ondřej and Pajdla, Tomáš and Sturm, Peter},  
year={2005},  month={Jan},  pages={86–102},  language={en-US}  }
}

\end{document}